\documentclass[lettersize,journal]{IEEEtran}
\usepackage{amsmath,amsfonts}
\usepackage{algorithmic}
\usepackage{algorithm}
\usepackage{array}
\usepackage[caption=false,font=normalsize,labelfont=sf,textfont=sf]{subfig}
\usepackage{textcomp}
\usepackage{stfloats}
\usepackage{url}
\usepackage{verbatim}
\usepackage{graphicx}
\usepackage{cite}
\usepackage{multirow}
\usepackage{bm} 
\usepackage{hyperref}
\usepackage{color}
\begin{document}

\title{Generalized Trusted Multi-view Classification Framework with Hierarchical Opinion Aggregation}

\author{Long Shi,~\IEEEmembership{Member, IEEE}, Chuanqing Tang, Huangyi Deng, Cai Xu, Lei Xing,~\IEEEmembership{Member, IEEE}, Badong Chen, ~\IEEEmembership{Senior Member, IEEE} 
\thanks{Long Shi, Chuanqing Tang and Huangyi Deng are with the School of Computing and Artificial Intelligence, and also with the Financial Intelligence and Financial Engineering Key Laboratory of Sichuan Province, Southwestern University of Finance and Economics, Chengdu 611130, China (e-mail: shilong@swufe.edu.cn, 223081200012@smail.swufe.edu.cn, 223081200043@smail.swufe.edu.cn)}
\thanks{Cai Xu is with the School of Computer Science and Technology, Xidian University, China (e-mail: cxu@xidian.edu.cn)}
\thanks{Lei Xing and Badong Chen are with the Institute of Artificial Intelligence and Robotics, Xi’an Jiaotong University, Xi’an 710049, China (e-mail: xingl@xjtu.edu.cn, chenbd@mail.xjtu.edu.cn)}
}

\markboth{Published as a journal paper at IEEE Transactions on Multimedia}%
{Shell \MakeLowercase{\textit{et al.}}: A Sample Article Using IEEEtran.cls for IEEE Journals}


\maketitle

\begin{abstract}
Recently, multi-view learning has witnessed a considerable interest on the research of trusted decision-making. Previous methods are mainly inspired from an important paper published by Han \emph{et al}. in 2021, which formulates a Trusted Multi-view Classification (TMC) framework that aggregates evidence from different views based on Dempster's combination rule. All these methods only consider inter-view aggregation, yet lacking exploitation of intra-view information. In this paper, we propose a generalized trusted multi-view classification framework with hierarchical opinion aggregation. This hierarchical framework includes a two-phase aggregation process: the intra-view and inter-view aggregation hierarchies. In the intra aggregation, we assume that each view is comprised of common information shared with other views, as well as its specific information. We then aggregate both the common and specific information. This aggregation phase is useful to eliminate the feature noise inherent to view itself, thereby improving the view quality. In the inter-view aggregation, we design an attention mechanism at the evidence level to facilitate opinion aggregation from different views. To the best of our knowledge, this is one of the pioneering efforts to formulate a hierarchical aggregation framework in the trusted multi-view learning domain. Extensive experiments show that our model outperforms some state-of-art trust-related baselines. One can access the source code on \href{https://github.com/lshi91/GTMC-HOA}{https://github.com/lshi91/GTMC-HOA}.   
\end{abstract}


\section{Introduction}
\IEEEPARstart{M}{ulti-view} learning has garnered increasing interest due to the prevalence of multi-source data in real-world scenarios. A typical example can be observed in the healthcare field, where a patient’s diagnosis is often made based on multiple types of medical examinations. According to the learning paradigm, multi-view learning methods can be broadly categorized into subspace-based methods \cite{gao2015multi, cao2023robust}, graph-based methods \cite{wang2019gmc, wen2020adaptive}, and deep neural network-based methods \cite{wang2020deep, yan2021deep}, which have found diverse applications in classification \cite{kan2016multi}, clustering \cite{wang2021fast}, recommendation systems \cite{lee2023mvfs}, among others.

Most multi-view methods follow one of two technical routes in their research. The first route argues that focusing on the shared information among views is sufficient to uncover the underlying data structure \cite{wang2015robust}. The second route seeks both consistent and complementary information \cite{luo2018consistent}. While both technical routes have significantly advanced the accuracy of decision-making, they often neglect the crucial aspect of result reliability, which is particularly important in safety-critical applications.  

\begin{figure}[htbp]
	\centering
	\includegraphics[scale=0.6]{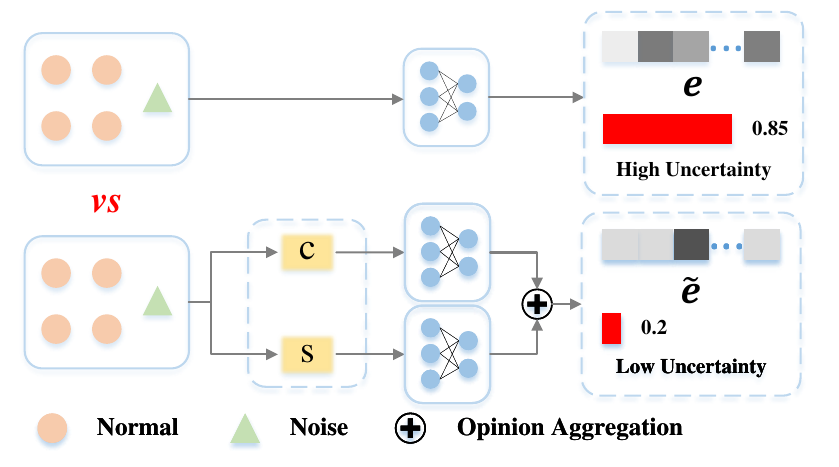} 
	\caption{For a noisy view, evidence directly from a neural network may have high uncertainty, while evidence after intra-view aggregation shows lower uncertainty, thus improving the reliability of view quality.}
	\label{motivationFig}
\end{figure} 

To address the aforementioned limitation, Han \emph{et al}. proposed an uncertainty-aware Trusted Multi-view Classification (TMC) framework \cite{han2022trusted}. TMC utilizes the Dirichlet distribution to model the distribution of class probabilities and incorporates the Dempster-Shafer theory \cite{shafer1976mathematical} to yield a reliable classification result. Inspired by TMC, several subsequent studies have been conducted, focusing on areas such as conflictive\footnotemark \footnotetext{In a multi-view scenario, data instance conflict arises when different views of the same instance contain contradictory or unrelated information.} opinion aggregation strategy \cite{xu2024reliable}, opinion aggregation with evidence accumulation \cite{liu2022trusted}, and trusted learning in the presence of label noise \cite{xu2024trusted}. Essentially, these methods can be characterized as following the paradigm of inter-view aggregation, they do not make efforts to investigate the potential benefits of exploiting intra-view information. But we argue that incorporating intra-view information is of significant importance. To explain it intuitively, we provide an example involving noisy multi-view data, as shown in Fig. \ref{motivationFig}. In such a scene, previous trusted methods produce large uncertainty in classification evidence, while resulting in limited classification accuracy. This observation motivates us to consider a problem: \emph{can we reduce the uncertainty of noisy view data to enhance view quality, thereby improving decision-making accuracy and reliability?}    

In response to the aforementioned challenge, we propose a novel trusted learning framework for multi-view data, namely Generalized TMC framework with Hierarchical Opinion Aggregation (GTMC-HOA). In the new framework, we propose a two-phase hierarchical aggregation process: intra-view aggregation and inter-view aggregation. The intra-view aggregation is realized by first learning the common information shared across different views, as well as the unique information specific to each view. Specifically, the common information is acquired by minimizing the confusion adversarial loss, while the view-specific information is obtained by enforcing an orthogonal constraint with respect to the common information. Once both the common and specific information for each view are obtained, we carry out the intra-view aggregation based on Dempster’s rule of combination. This aggregation phase is useful to reduce the uncertainty of evidence in the presence of noisy view, thereby enhancing the quality of each view. Subsequently, the improved views are used to facilitate inter-view aggregation. This aggregation phase involves two procedures. Firstly, an attention mechanism is designed at the level of evidence to extract attention from other views, which circumvents the limitation of previous trusted methods that treat each view equally, thereby further reinforcing evidence reliability. Secondly, we perform the inter-view aggregation. In summary, \emph{we are the first to propose a hierarchical aggregation framework in trusted multi-view learning}, and our main contributions of GTMC-HOA include:

\begin{itemize}
    \item In the first hierarchy, the intra-view aggregation, we learn both the common and specific components for each view, and then conduct intra-view aggregation. This hierarchy of aggregation stands from the aspect of intra views, which helps to reduce evidence uncertainty, particularly in the presence of noisy data.     
    \item In the second hierarchy, the inter-view aggregation, we first design an attention mechanism at the evidence level to capture information from other views, and then execute inter-view aggregation. This hierarchy of aggregation is constructed from the perspective of inter views, enhancing the quality of evidence and ultimately bolstering decision-making reliability.     
    \item Extensive experiments demonstrate that GTMC-HOA achieves more superior performance than several state-of-art trusted methods. 
\end{itemize}

\section{Related Work}
In the following, we provide a brief review of representative research in multi-view learning and uncertainty-aware trusted learning.  


\subsection{Multi-view Learning} Multi-view learning has seen rapid development in recent years. Multi-view subspace clustering methods receive particular interest because they allow analysis and understanding of data structures in the underlying low-dimensional subspace \cite{shi2024nonlinear, luo2018consistent,brbic2018multi}. Unlike traditional subspace clustering methods that learn self-representation from original features, some latent representation methods were investigated to more effectively capture comprehensive information among multiple views \cite{zhang2018generalized, chen2020multi}. The basic idea behind these methods is to project the original features into a latent space, and then learn a latent representation that covers both common and complementary information \cite{shi2024enhanced}. Due to the merits of kernel mapping in handling nonlinear structures \cite{wu2024low}, extensive research has been conducted on kernel-based multi-view methods. Liu \emph{et al} investigated the issue of incomplete kernel by unifying the procedures of kernel imputation and clustering \cite{liu2019multiple}. By considering the local density of samples, kernel representation capacity can be improved \cite{liu2020optimal}. In addition, tensor-based multi-view subspace learning that is capable of exploiting high-order information demonstrates superiority in capturing the correlations across various views \cite{xie2018unifying, wu2020unified, qin2023flexible}. 


Graph techniques also occupy an important position in multi-view learning. Considerable efforts have been devoted to graph-based multi-view clustering \cite{wang2019gmc}, addressing specific issues such as consensus graph learning \cite{zhan2018multiview, li2021consensus, wang2022towards}, graph recovering \cite{wong2019clustering, wen2020adaptive}, and attributed graphs \cite{lin2021multi, lin2021graph}. With the rapid advancement of deep learning techniques, there has been growing interest in integrating multi-view learning with deep neural networks \cite{chen2023deep, lu2024decoupled, zhu2024multiview}. Moreover, a variety of practical challenges have been actively explored in the literature, including--but not limited to--low-quality views \cite{wen2020adaptive, liu2023incomplete}, particularly the challenge of handling incomplete multi-view data \cite{zhu2022latent,wang2024integrated}, and the efficient processing of large-scale data \cite{wang2021fast, yang2024fast}.

\subsection{Uncertainty-aware Trusted Learning} However, the aforementioned multi-view learning methods only focus on decision-making accuracy, neglecting the crucial aspect of result reliability, which is important for safety-critical applications. Motivated by evidential deep learning \cite{sensoy2018evidential} that serves as a representative paradigm for uncertainty estimation, TMC has recently emerged as a significant framework, which utilizes the Dirichlet distribution to model class probability distributions and incorporate the Dempster-Shafer theory to realize trusted classification results \cite{han2020trusted, han2022trusted}. On the basis of TMC, various variants have been developed to address specific issues and enhance performance. Liu \emph{et al}. considered opinion entropy to ensure consistency among multiple views \cite{liu2022trusted}. Xu \emph{et al}. investigated the reliable conflictive multi-view learning problem by designing a conflictive opinion aggregation strategy \cite{xu2024reliable}. Subsequently, they explored schemes to guarantee the reliability of decision-making in scenarios involving noisy labels \cite{xu2024trusted} and semantic vagueness \cite{liu2024dynamic}. The potential of incorporating trustworthiness in both early and late fusion stages has been investigated by Zhou \emph{et al}. \cite{zhou2023calm}, who demonstrated its effectiveness in enhancing prediction quality. More advancements on trusted learning can be found in recent work such as \cite{zhou2023rtmc, wang2024trusted, du2023bridging, zou2023dpnet}, which covers categories including semi-supervised learning, multimodal fusion, and open-world learning, among others.

\section{Methodology}
\begin{figure*}[htbp]
	\centering
	\includegraphics[scale=0.3]{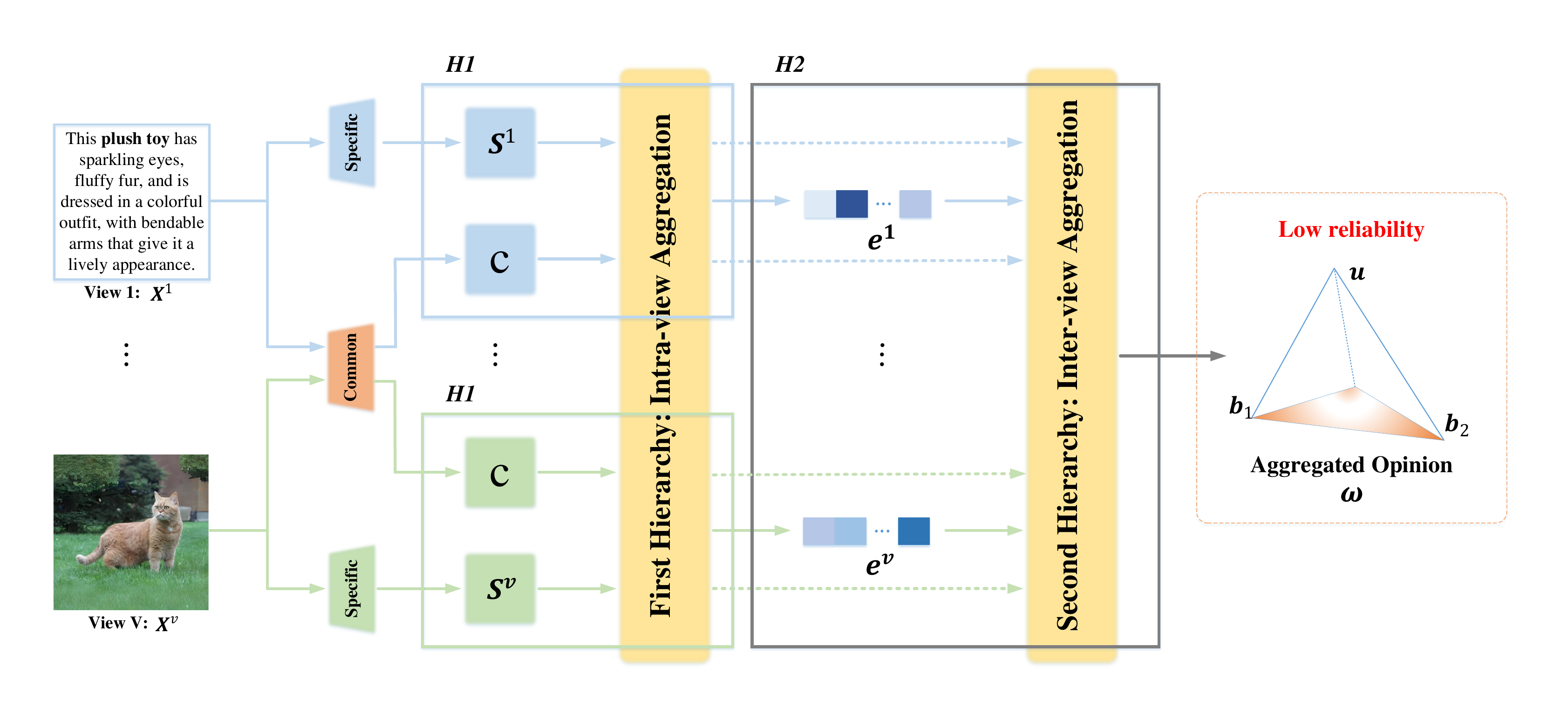} 
	\caption{Illustration of GTMC-HOA. First, we learn the common and specific information for each view. Then, we perform the first-level aggregation, namely intra-view aggregation, to generate high-quality evidence. After obtaining the high-quality evidence of multiple views, we proceed with the second-level aggregation, namely inter-view aggregation, to achieve reliable decision-making results.}
	\label{mainFramework}
\end{figure*}

In this section, we introduce our proposed GTMC-HOA in detail, which involves three modules: learning of common and specific information, intra-view aggregation, and inter-view aggregation. We also provide a theoretical analysis of the computational complexity of the proposed model. The framework of GTMC-HOA is shown in Fig. \ref{mainFramework}. In this framework, intra-view aggregation is depicted in the \emph{H1} block, while inter-view aggregation is represented in the \emph{H2} block. For a $q$-class classification problem, we consider multi-view data $\{\boldsymbol{X}^{i}\}^v_{i=1}$ with $n$ samples, where $\boldsymbol{X}^{i} = [\boldsymbol{x}^i_1, \boldsymbol{x}^i_2,\cdots, \boldsymbol{x}^i_n] \in \mathbb{R}^{d_i\times n}$ represents the $i$-th view data with $d_i$ dimension.

\subsection{Common and Specific Information}

Before implementing the intra-view aggregation, we need to learn both the common and specific information for each view. 

\noindent \textbf{Extraction of Common Information}: Motivated by \cite{liu2017adversarial}, we extract the common information by minimizing an adversarial loss $\mathcal{L}_{adv}$, which serves to confuse the discriminator as to which view the learned common information belongs. 

Consider $\boldsymbol{c}^i_j$ as the common subspace representation of the original $i$-th view feature $\boldsymbol{x}^i_j$ from the $j$-th sample. This representation is extracted using a common subspace extraction layer $H_{cse}$, such that $\boldsymbol{c}^i_j = H_{cse}(\phi^i(\boldsymbol{x}^i_j))$, where $\phi^i(\cdot)$ maps each $d_i$-dimensional original feature vector into the same $l$-dimensional space. Define $\boldsymbol{z}_j$ as the $v$-dimensional view label vector of $\boldsymbol{c}_j^i$, where $\boldsymbol{z}_j^i$ equals $1$ and all other elements are $0$, indicating that $\boldsymbol{c}_j^i$ originates from the $i$-th view. A training set $\mathcal{T}_{adv}=\{ (\boldsymbol{c}_j^i,\boldsymbol{z}_j)|1 \leq i \leq v,1 \leq j \leq n \}$ can be constructed for the discriminator $D$. The prediction output by $D$ is denoted as $\hat{\boldsymbol{z}}$, such that $\hat{\boldsymbol{z}}_j=D(\boldsymbol{c}_j^i)$. The adversarial loss $\mathcal{L}_{adv}$ is then defined as:
\begin{equation}
\label{eq01}
    \mathcal{L}_{adv}=\mathcal{F} \left( -\sum_{j=1}^n \sum_{i=1}^v {z}_j^i \log(\hat{{z}}_j^i) \right),
\end{equation}
where $\mathcal{F}(\cdot)$ is a monotonically decreasing function set to $\mathcal{F}(x)=\exp(-x)$. With this loss function, we are able to confuse the discriminator, preventing it from identifying the true view of the learned subspace representation. Consequently, the incorporated representation $\boldsymbol{c}^i_j$ only contains the common information from $\boldsymbol{x}^i_j$.

To avoid the unexpected situation that noise may confuse the discriminator, we draw inspiration from \cite{wu2019multi} and introduce a common subspace multi-view loss $\mathcal{L}_{cml}$ to ensure that $\boldsymbol{c}^i$ contains certain semantics. Consider a training set $\mathcal{T}_{cml}=\{ (\boldsymbol{c}_j^i,\boldsymbol{y}_j)|1 \leq i \leq v,1 \leq j \leq n \}$ for common subspace representation prediction layer $H_{cml}$. Let $\hat{\boldsymbol{y}}$ represent the output by $H_{cml}$, such that $\hat{\boldsymbol{y}}_j^i=H_{cml}(\boldsymbol{c}_j^i)$. The loss $\mathcal{L}_{cml}$ can then be defined as:      
\begin{equation}
\label{eq02}
    \mathcal{L}_{cml}=-\sum_{j=1}^n \sum_{i=1}^v \sum_{k=1}^q {y}_{jk} 
    \log(\hat{{y}}_{jk}^i) + (1-{y}_{jk}) \log(1-\hat{{y}}_{jk}^i)).  
\end{equation}

By incorporating $\mathcal{L}_{adv}$ and $\mathcal{L}_{cml}$, the overall loss $\mathcal{L}_{com}$ for learning  common information can be expressed as:
\begin{equation}
\label{eq03}
    \mathcal{L}_{com}=\mathcal{L}_{adv} + \mathcal{L}_{cml}.
\end{equation}

\noindent \textbf{Extraction of Specific Information}: In an ideal situation, for each view, we aim for the learned common information to be independent of the learned specific information. To achieve this, we enforce an orthogonal constraint on the specific information. Consider $\boldsymbol{s}^i$ as the $l$-dimensional feature vector extracted by specific information extraction layer $H_{sie}$, such that $\boldsymbol{s}^i=H_{sie}(\boldsymbol{x}^i)$.  We then define the following constraint to encourage orthogonality between $\boldsymbol{s}^i$ and $\boldsymbol{c}$:  
\begin{equation}
\label{eq04}
    \mathcal{L}_{spe} = \left\| (\boldsymbol{s}^{i})^T \boldsymbol{c} \right\|_2^2,
\end{equation}
where $\boldsymbol{c}=\sum^v_{i=1}\boldsymbol{c}^{i}$, and $\| \cdot \|_2$ denotes the $l_2$-norm. From the above equation, it is seen that $\boldsymbol{s}^{i}$ is orthogonal to $\{\boldsymbol{c}^1, \cdots, \boldsymbol{c}^v\}$, ensuring that the specific information is as different from the common information as possible.

\subsection{Intra-view Aggregation}

\begin{figure*}[htbp]
	\centering
	\includegraphics[scale=0.21]{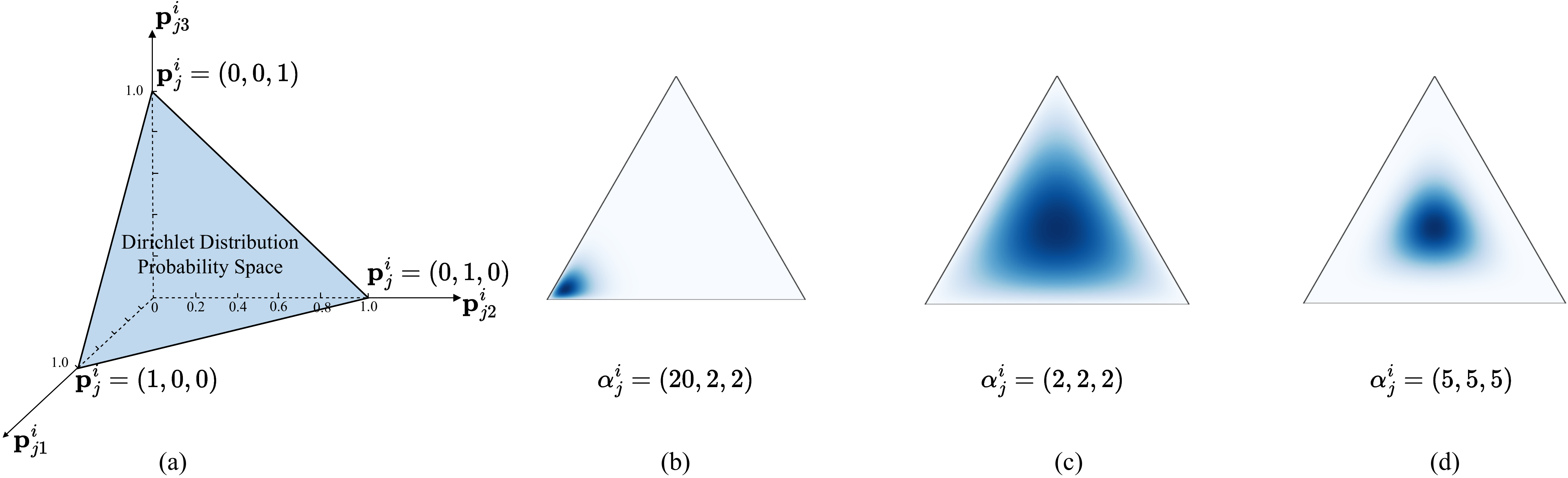} 
	\caption{Examples of Dirichlet distribution and subjective opinion. (a) Possibility space (represented as a standard 2-simplex) for the probability of categories in a three-classification problem. (b)-(d) Prediction cases with different confidence, corresponding to evidence $\mathbf{e}^i_j=\{19,1,1\}$, $\mathbf{e}^i_j=\{1,1,1\}$, and $\mathbf{e}^i_j=\{4,4,4\}$, respectively.}
	\label{Dirichlet_distribution}
\end{figure*} 

\begin{figure}[htbp]
	\centering
	\includegraphics[scale=0.32]{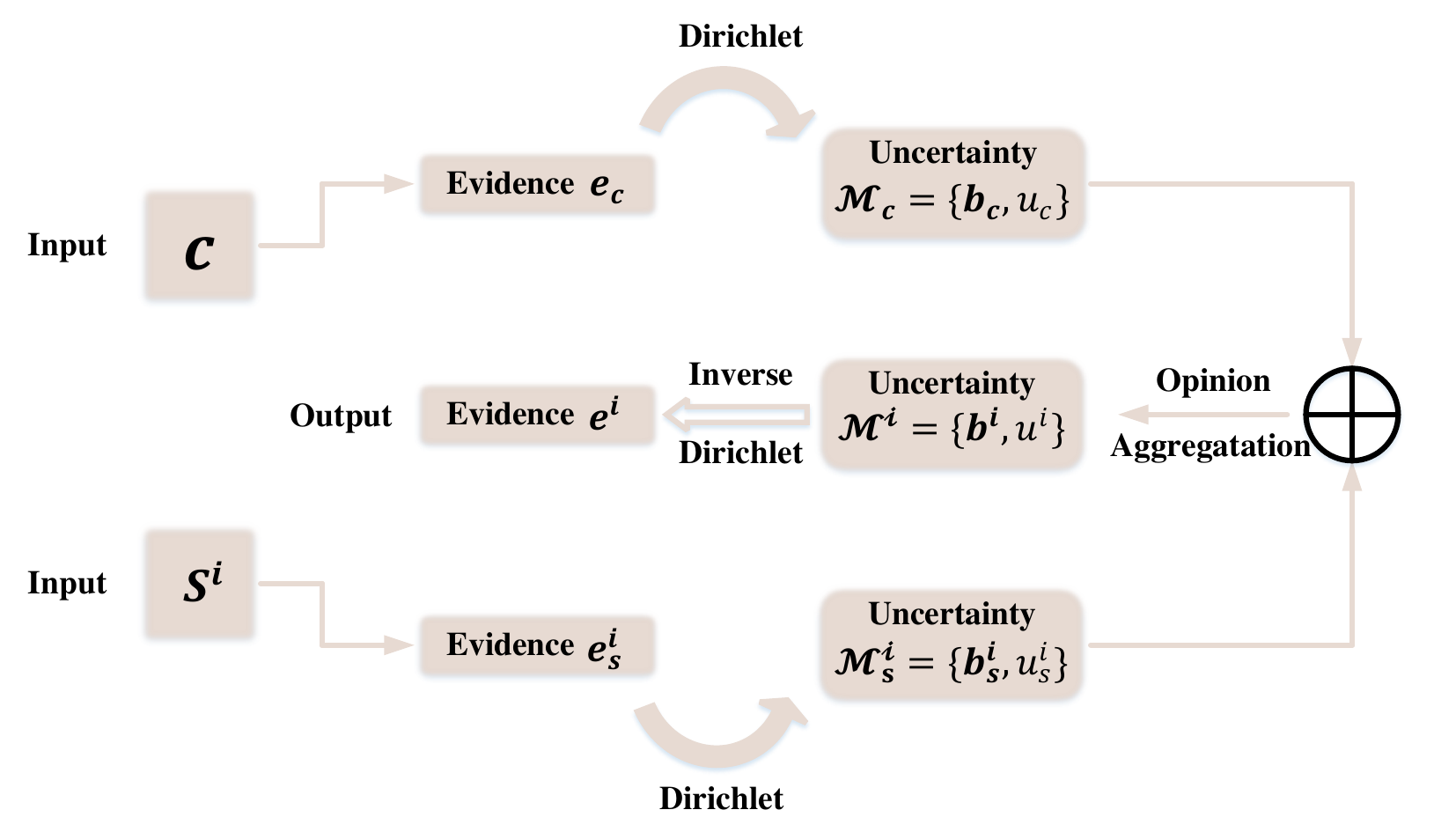} 
	\caption{Structure of intra-view aggregation. The common and specific information of each view are aggregated to reduce the inherent uncertainty.}
	\label{intraviewAggregation}
\end{figure}

After obtaining both common and specific information for each view, we execute the first phase of aggregation, namely intra-view aggregation, as shown in Fig. \ref{intraviewAggregation}. During this phase, we first learn evidence corresponding to the common and specific information by deep neural networks (DNNs), where ``evidence" represents metrics collected from the input to support the classification. The theoretical framework of subjective logic (SL) \cite{jsang2018subjective} is then employed to associate the parameters of the Dirichlet distribution based on the gathered evidence, thereby obtaining the probabilities of different classes and the uncertainty of decision-making. Finally, opinion aggregation is performed.   

\noindent \textbf{Evidential Deep Learning}: Most deep learning methods utilize a softmax layer for the predictive output, but they fail to estimate predictive uncertainty. In contrast, evidential deep learning, which incorporates uncertainty estimation, can be a preferred alternative. For a $q$-class classification problem, SL aims to assign a belief mass to each class label and an overall uncertainty mass to the whole framework, which, for the $i$-th view, is formulated as the following equation with $q+1$ mass values:
\begin{equation}  
\label{eq05}
u^i + \sum_{k=1}^{q} b_k^i = 1,
\end{equation}
where $u^i$ and $b_k^i$ denote the overall uncertainty and the probability for the $k$-th class, respectively, and they are all non-negative.

SL defines the relationship between the evidence $\boldsymbol{e}^i = [e^i_1,\cdots,e_q^i]$ and the parameters of the Dirichlet distribution $\boldsymbol{\alpha}^i = [\alpha^i_1,\cdots,\alpha_q^i]$. Specifically, the parameter $\alpha^i_k$ of the Dirichlet distribution is induced from $e^i_k$, such that $\alpha^i_k = e^i_k + 1$. Then, the belief mass $b^i_k$ and the uncertainty $u^i$ are calculated by
\begin{equation} 
\label{eq06} 
b_k^i = \frac{e_k^i}{S^i} = \frac{\alpha_k^i - 1}{S^i} \quad \text{and} \quad u^i = \frac{q}{S^i},
\end{equation}
where $S^i = \sum\nolimits_{k=1}^{q} (e_k^i + 1) = \sum\nolimits_{k=1}^{q} \alpha^i_k$ denotes the Dirichlet strength. Eq. (\ref{eq06}) reflects two observations: First, the more evidence gathered for the $k$-th category, the higher the probability assigned to the $k$-th class. Second, the less total evidence observed, the greater the overall uncertainty. 

To better illustrate the concepts of evidence theory, let's delve into a three-class classification example. Fig. \ref{Dirichlet_distribution}(a) represents the probability space (standard 2-simplex) for a three-class problem. Consider a high-confidence classification case where the network collects evidence $\boldsymbol{e}_j^i = \{19, 1, 1\}$, aligning with a Dirichlet distribution having parameters $\boldsymbol{\alpha}_j^i = \{20, 2, 2\}$. Here, the uncertainty mass $u \approx 0.13$. As seen in Fig. \ref{Dirichlet_distribution}(b), this results in a distribution pattern concentrated near one vertex of the triangle. Alternatively, consider the case where the evidence for each class is relatively small, such that $\boldsymbol{e}_j^i = \{1, 1, 1\}$. This results in parameters $\boldsymbol{\alpha}_j^i = \{2, 2, 2\}$ and an uncertainty mass $u \approx 0.5$. Fig. \ref{Dirichlet_distribution}(c) illustrates how this evidence leads to a more dispersed distribution across the simplex. We also consider the case where $\boldsymbol{e}_j^i = \{4, 4, 4\}$, which corresponds to Dirichlet distribution parameters $\boldsymbol{\alpha}_j^i = \{5, 5, 5\}$ and an uncertainty mass of $u \approx 0.2$. As shown in Fig. \ref{Dirichlet_distribution}(d), compared to the previous cases, this distribution is more focused towards the center of the simplex, indicating that an increase in evidence amount leads to reduced overall uncertainty.

\noindent \textbf{Opinion Aggregation}: The result obtained from Eq. (\ref{eq06}) can be considered a subjective opinion. For each view, we have two opinions for common and specific information, denoted as $\boldsymbol{\omega}_{c}=\{\{b_{c,k}\}^q_{k=1}, u_{c}\}$ and $\boldsymbol{\omega}_{s}=\{\{b_{s,k}\}^q_{k=1}, u_{s}\}$, respectively. These two opinions need to be aggregated. Previous works have presented several advancements in opinion aggregation. In this paper, we select the latest state-of-the-art aggregation method, as reported in \cite{xu2024reliable}, which enables conflicting opinion aggregation. Readers are encouraged to refer to the corresponding literature for details.  Upon obtaining the aggregated opinion $\boldsymbol{\omega}=\{\{b_{k}\}^q_{k=1}, u\}$, we can inversely recover the corresponding evidence.  

\textbf{\emph{Remark}}: By learning both common and specific information, we capture a more comprehensive representation of the view. Evidence theory allows us to model the uncertainty accociated with each component of information, enabling an enhanced fusion that prioritizes reliable data while mitigating the impact of noise.

\subsection{Inter-view Aggregation}

\begin{figure*}[htbp]
	\centering
	\includegraphics[scale=0.5]{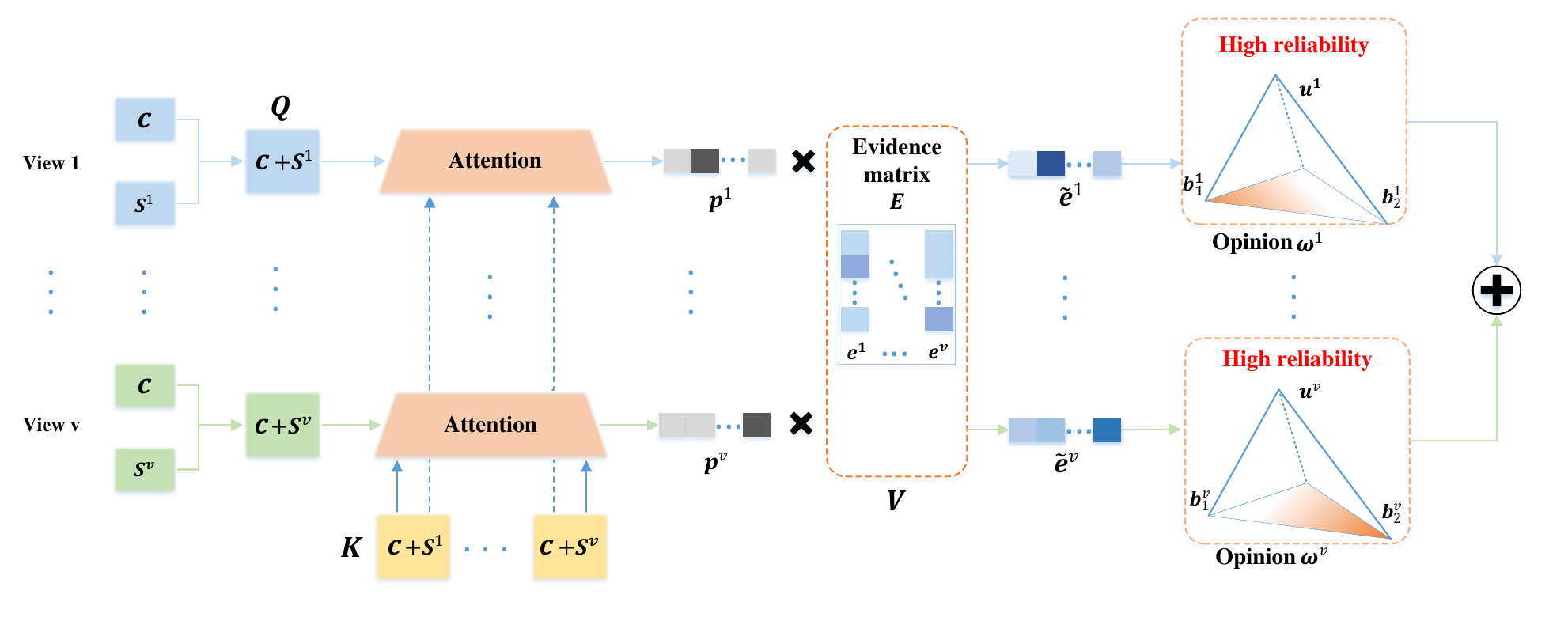} 
	\caption{Structure of inter-view aggregation. An attention mechanism at the evidence level is designed to enhance the importance of high-quality evidence.}
	\label{interviewAggregation}
\end{figure*}

Existing trusted multi-view learning methods treat all views equally, leading to suboptimal opinion aggregation among views. This can result in the model being less effective at distinguishing between more and less relevant views, potentially weakening the impact of important views and their corresponding evidence, thereby reducing overall performance. To address this, we incorporate an attention mechanism for inter-view aggregation to capture importance scores from other views, enhancing the evidence from high-quality views. The implementation procedures of inter-view aggregation is shown in Fig. \ref{interviewAggregation}. 


Since we have learned both common and specific information for each view, it is more effective to combine these to represent the view information. Thus, the $i$-th view is represented by a common feature $\boldsymbol{c}$ and a view-specific feature $\boldsymbol{s}^i$, denoted as $\boldsymbol{c}+\boldsymbol{s}^i$ with size of $l\times 1$. The corresponding query $\boldsymbol{Q}$ is then defined by:
\begin{equation}
\label{eq07}
    \boldsymbol{Q} = \boldsymbol{W}^Q \boldsymbol{F}_{cs},
\end{equation}
where $\boldsymbol{W}^Q\in \mathbb{R}^{v\times v}$ denotes the query weight matrix, and $\boldsymbol{F}_{cs} = \left[\boldsymbol{c} +\boldsymbol{s}^1,\boldsymbol{c} +\boldsymbol{s}^2,\dots,\boldsymbol{c} +\boldsymbol{s}^v \right]^T \in \mathbb{R}^{v\times l}$. We then define the update for the key $\boldsymbol{K}$ as follows:
\begin{equation}
\label{eq08}
    \boldsymbol{K} =  \boldsymbol{W}^K \boldsymbol{F}_{cs},
\end{equation}      
where $\boldsymbol{W}^K\in \mathbb{R}^{v\times v}$ is the key weight matrix. Given that our purpose is to dynamically select the most favorable evidence for classification, the value $\boldsymbol{V}$ should be evidence-related and is defined as: 
\begin{equation}
\label{eq09}
    \boldsymbol{V} = \boldsymbol{W}^V \boldsymbol{E},
\end{equation}   
where $\boldsymbol{W}^V\in \mathbb{R}^{v\times v}$ is the value weight matrix, and $\boldsymbol{E} = [\boldsymbol{e}^1, \boldsymbol{e}^2, \cdots, \boldsymbol{e}^v]^T\in \mathbb{R}^{v\times q}$ denotes the evidence matrix.



Conventional attention mechanisms use a softmax function for dynamic weight selection. However, applying softmax in this context can cause the network to overly rely on evidence from a single view, as the weights for other views may approach zero. This leads to overconfidence in one view and hinders the effective utilization of information from all views. To address this, we adopt the ReLU activation function to achieve more balanced and reasonable weight selection. With attention considered, we can calculate the evidence for the $i$-th view by 


\begin{equation}
\label{eq10}
{\rm{Attention}}(\boldsymbol{Q}^i,\boldsymbol{K},\boldsymbol{V}) = \frac{{\rm{ReLU}} \left( \boldsymbol{Q}^i \boldsymbol{K}^T / \sqrt{l}  \right) + \epsilon}{{\rm{sum}} \left({\rm{{ReLU}}} \left( \boldsymbol{Q}^i \boldsymbol{K}^T / {\sqrt{l}} \right) + \epsilon \right)} \boldsymbol{V},
\end{equation}    
where $\boldsymbol{Q}^i$ represents the $i$-th row of $\boldsymbol{Q}$, and $\epsilon$ is a small value added to prevent division by zero.

After obtaining the attention scores for each view, the importance of evidence from high-quality views is enhanced, while that from less significant views is reduced. This approach allows the model to focus more on critical evidence that significantly influences decision-making. Subsequently, we perform opinion aggregation between views using the aforementioned conflicting opinion strategy.

\subsection{Loss Function}

The overall loss function is composed of three parts: extraction of common and specific information, intra-view aggregation, and inter-view aggregation. In the earlier subsection, we have introduced the loss functions for learning common and specific information, namely $\mathcal{L}_{com}$ and $\mathcal{L}_{spe}$. Therefore, in this section, we focus on the details of the loss functions for intra-view and inter-view aggregation.  

Before proceeding, we need to introduce some fundamental concepts related to designing loss functions in evidential deep learning. In this context, the softmax layer of a neural network is replaced with an activation function layer (i.e., ReLU) to ensure that the network outputs non-negative values, which are interpreted as the evidence vector. When designing the classification loss in trusted multi-view learning, the traditional cross-entropy loss must be adapted to align with the evidence-based approach in the case of $j$-th instance for $k$-th class:
\begin{equation} 
\label{eq11}
    \begin{aligned}  
        \mathcal{L}_{ace}(\boldsymbol{\alpha}_j) &= \int \left[ \sum_{k=1}^{q}-y_{jk}\log p_{jk} \right] \frac{\bm{\Pi}_{k=1}^q p_{jk}^{\alpha_{jk}-1}}{B(\boldsymbol{\alpha}_j)}d{\boldsymbol{p}_j}\\   
        &= \sum_{k=1}^{q} y_{jk}(\psi(S_j) - \psi(\alpha_{jk})),
    \end{aligned}  
\end{equation}
where $\boldsymbol{\alpha}_j$ denotes the parameter vector (i.e., $\boldsymbol{\alpha}^i_j = \boldsymbol{e}^i_j + \boldsymbol{1}$) of the Dirichlet distribution, and $\psi(\cdot)$ represents the digamma function. The above loss function ensures that the correct label of each sample generates more evidence than other classes. However, it does not guarantee that less evidence will be generated for incorrect labels. To compensate this drawback, an additional Kullback-Leibler (KL) divergence term is required:
\begin{equation}
\label{eq12}
    \begin{aligned} 
        \mathcal{L}_{KL}(\boldsymbol{\alpha}_j) &= KL \left[D(\boldsymbol{p}_j|\tilde{\boldsymbol{\alpha}}_j)||D(\boldsymbol{p}_j|\boldsymbol{1})\right]\\
        &= \log(\frac{\Gamma(\sum_{k=1}^{q} \tilde{\alpha}_{jk})}{\Gamma(q)\Pi^q_{k=1}\Gamma(\tilde{\alpha}_{jk})})\\
        &+\sum_{k=1}^{q}(\tilde{\alpha}_{jk}-1) \left[\psi(\tilde{\alpha}_{jk}) - \psi(\sum_{m=1}^{q}\tilde{\alpha}_{jm})\right],
    \end{aligned}  
\end{equation}
where $D(\boldsymbol{p}_j|\boldsymbol{1})$ is the uniform Dirichlet distribution,$\tilde{\boldsymbol{\alpha}}_{j} = y_j + (1-y_j)\odot \boldsymbol{\alpha}_j$ is the Dirichlet parameter that prevents the evidence for the ground truth class from being penalized to zero, and $\Gamma(\cdot)$ stands for gamma function. Combining (\ref{eq11}) and (\ref{eq12}), we have
\begin{equation}
\label{eq13}
    \mathcal{L}_{acc}(\boldsymbol{\alpha}_j) = \mathcal{L}_{ace}(\boldsymbol{\alpha}_j) + \lambda_t \mathcal{L}_{KL}(\boldsymbol{\alpha}_j),
\end{equation}
where $\lambda_t$ is a positive balance factor with $t$ being the current training epoch index.  

In the intra-view aggregation (i.e., \emph{H1} hierarchy), apart from the basic classification loss $\mathcal{L}_{ace}(\boldsymbol{\alpha}^i_j)$, we also hope that the extracted common and specific information can guide correct classification. To achieve this, we incorporate cross-entropy losses for the common and specific information, denoted as $\mathcal{L}_{ace}(\boldsymbol{\alpha}_{c,j})$ and $\mathcal{L}_{ace}(\boldsymbol{\alpha}^{i}_{s,j}) $, respectively. Since our model employs a conflictive opinion aggregation strategy, there is an additional loss term for the degree of conflict between common and specific components. Therefore, the loss for \emph{H1} is given by
\begin{equation}
\begin{aligned}
\label{eq14}
    \mathcal{L}_{H1} = &\frac{1}{v} \sum_{i=1}^v \left[  
    \mathcal{L}_{ace} (\boldsymbol{\alpha}_j^i) + \left(\mathcal{L}_{ace}(\boldsymbol{\alpha}_{c,j})+\mathcal{L}_{ace}(\boldsymbol{\alpha}^{i}_{s,j})  \right) \right.
    \\
    & \left. + \gamma C(\boldsymbol{\omega}_{c,j},\boldsymbol{\omega}^{i}_{s,j}) 
    \right],
\end{aligned}
\end{equation}
where $C(\boldsymbol{\omega}_{c,j},\boldsymbol{\omega}^{i}_{s,j})$ characterizes the degree of conflict between common and specific components with the corresponding calculation detailed in \cite{xu2024reliable}, and $\gamma$ is a trade-off parameter.  
        
In the inter-view aggregation (i.e., \emph{H2} hierarchy), we use (\ref{eq13}) for the basic classification loss. Additionally, we aim for the attention-induced views to achieve reliable classification, so we introduce an extra classification loss for these views. Furthermore, we account for the conflict degree loss between views. Thus, we have 
\begin{equation}
\label{eq15}
    \mathcal{L}_{H2} = \mathcal{L}_{acc}(\boldsymbol{\alpha}_j) + \sum_{i=1}^{v}\mathcal{L}_{acc}(\boldsymbol{\hat{\alpha}}_j^i) + \gamma \mathcal{L}_{con},
\end{equation}
where $\boldsymbol{\hat{\alpha}}_j$ is the parameter vector of the Dirichlet distribution after applying attention, and $\mathcal{L}_{con}$ is given by
\begin{equation}
\label{eq16}
    \mathcal{L}_{con} = \frac{1}{v-1}\sum_{p=1}^{v} \left( \sum_{q \neq p}^{v}C(\boldsymbol{\omega}^p_j,\boldsymbol{\omega}^q_j) \right).
\end{equation}

In summary, the overall loss of our model is shown as:
 
\begin{equation}
\label{eq17}
    \mathcal{L}_{overall} = \mathcal{L}_{H1} + \mathcal{L}_{H2} + \delta \mathcal{L}_{com} + \eta \mathcal{L}_{spe},
\end{equation}
where $\delta$ and $\eta$ are trade-off parameters. The terms in the loss function enable us to balance consistency and complementarity through the parameters $\delta$ and $\eta$. Considering the dominant role of common information, we tend to assign a larger value to $\delta$ compared to $\eta$. The implementation pseudocode of our model is summarized in \textbf{Algorithm \ref{algorithm1}}.

\captionsetup[algorithm]{labelfont=bf,labelsep=space}
\begin{algorithm}
    \caption{Implementation pseudocode of GTMC-HOA}
    \begin{algorithmic}
        \item[]\hspace*{-\algorithmicindent}\textbf{Input:} Multi-view dataset: $\{\boldsymbol{X}^i, \boldsymbol{y}\}_{i=1}^v$
        \item[]\hspace*{-\algorithmicindent}\textbf{Parameter:} Hyper-parameter: $l=64, \gamma=1, \delta=1,\eta=\{0.1,0.01,0.001\}$
         \item[]\hspace*{-\algorithmicindent}/---Train---/
         \item[]\hspace*{-\algorithmicindent}\textbf{Output:} Network's parameters.
        \STATE 01: \textbf{while} not converged \textbf{do}
        \STATE 02: \quad Initialize common features $\boldsymbol{c}$.
        \STATE 03: \quad \textbf{for} $i=1:v$  \textbf{do}
        \STATE 04: \qquad $\boldsymbol{c}^i \leftarrow$ common subspace network output.
        \STATE 05: \qquad $\boldsymbol{s}^i \leftarrow$ specific information network output.
        \STATE 06: \qquad Calculate $\boldsymbol{c}=\boldsymbol{c}+\boldsymbol{c}^i$.
        \STATE 07: \quad Calculate common information $\boldsymbol{c}=\boldsymbol{c}/v$.
        \STATE 08: \quad \textbf{for} $i=1:v$  \textbf{do}
        \STATE 09: \qquad $\boldsymbol{e}_c,\boldsymbol{e}_s^i \leftarrow$ evidential network output.
        \STATE 10: \qquad Calculate $\boldsymbol{e}^i$ by Intra-view Aggregation.
        \STATE 11: \quad \textbf{for} $i=1:v$  \textbf{do}
        \STATE 12: \qquad Calculate $\boldsymbol{\hat{e}}^i$ by Inter-view Aggregation.
        \STATE 13: \qquad Calculate distribution $\boldsymbol{\hat{\alpha}}^i=\boldsymbol{\hat{e}}^i+1$.
        \STATE 14: \quad Calculate the joint distribution $\boldsymbol{\alpha}$. 
        \STATE 15: \quad Calculate the overall loss $\mathcal{L}_{overall}$ by Eq. (17).
        \STATE 16: \quad Update the networks by gradient descent.
        \STATE 17: \textbf{end while}
        \STATE 18: \textbf{return} networks parameters. 
        \item[]\hspace*{-\algorithmicindent}/---Test---/
        \item[]\hspace*{-\algorithmicindent}\textbf{Output:}The final prediction and corresponding uncertainty. 
        \STATE 01: Repeat training steps 02-14.
        \STATE 02: Calculate the joint $\boldsymbol{b}, u$ by Eq. 6.
        \STATE 03: \textbf{return} the decision $\boldsymbol{b}$ and associated uncertainty $u$.
    \end{algorithmic}
\label{algorithm1}
\end{algorithm}

\subsection{Theoretical Complexity Analysis}

To provide a theoretical analysis of the model's computational complexity, we focus on the most significant components that dominate the computational cost, while disregarding minor contributions that do not substantially affect the overall complexity. 

The model's architecture consists of several key components, with the primary computational cost attributed to the view-specific encoders and the shared encoder. These components are built upon multi-layer fully connected networks that process input data from multiple views. Let $D_v=\sum^v_{i=1}d_i$ denote the total dimensionality across all views, $D_r$ represent the dimensionality of the learned representation, and $L$ be the number of layers (assuming both the shared and view-specific encoders have the same depth and representation dimensionality). The computational complexity for learning both common and view-specific representations can then be approximated as $O(nvD_vD_zL)$. Additionally, the inter-view aggregation incorporates an attention mechanism whose complexity is proportional to $O(v^2L)$. Therefore, the overall computational complexity of the model is $O(nvD_vD_zL+V^2L)$. Given that the number of samples $n$ is typically much larger than the number of view number $v$, the dominant terms becomes $O(nvD_vD_zL)$,which primarily governs the model's computational cost. In addition, we validate practical metrics, including the number of learnable parameters, MACs, and FLOPs, in the experimental section. 

\section{Experiments}

In this section, we compare the classification accuracy of GTMC-HOA with existing state-of-the-art methods, and investigate the model robustness against varying noise. We conduct an ablation study to figure out the contributions of main modules. Moreover, we analyze the conflictive degree, as well as the local and overall uncertainty on conflictive multi-view data. We also validate the practical complexity of our model.

\subsection{Experimental Setup}


\subsubsection{Datasets} 

\begin{table}[h!]
\caption{Dataset summary.}
\centering
\small
\begin{tabular}{|c|c|c|c|}
\hline
Dataset & Size & \textbf{$K$} & Dimensionality \\ \hline
HandWritten & 2000 & 10 & 240/76/216/47/64/6 \\ \hline
CUB & 600 & 10 & 1024/300 \\ \hline
Scene15 & 4485 & 15 & 20/59/40 \\ \hline
PIE & 680 & 68 & 484/256/279 \\ \hline
Caltech101 & 2386 & 20 & 48/40/254/1984/512/928 \\ \hline
NUS-WIDE & 2400 & 12 & 64/144/73/128/225/500 \\ \hline
NoisyMNIST & 20000 & 10 & 784/784 \\ \hline
Event8 & 1579 & 8 & 1000/1000/1000 \\ \hline
CCV & 6773 & 20 & 20/20/20 \\ \hline 
\end{tabular}
\label{Dataset_summary}
\end{table}

\begin{figure}[htbp]
	\centering
	\includegraphics[scale=0.2]{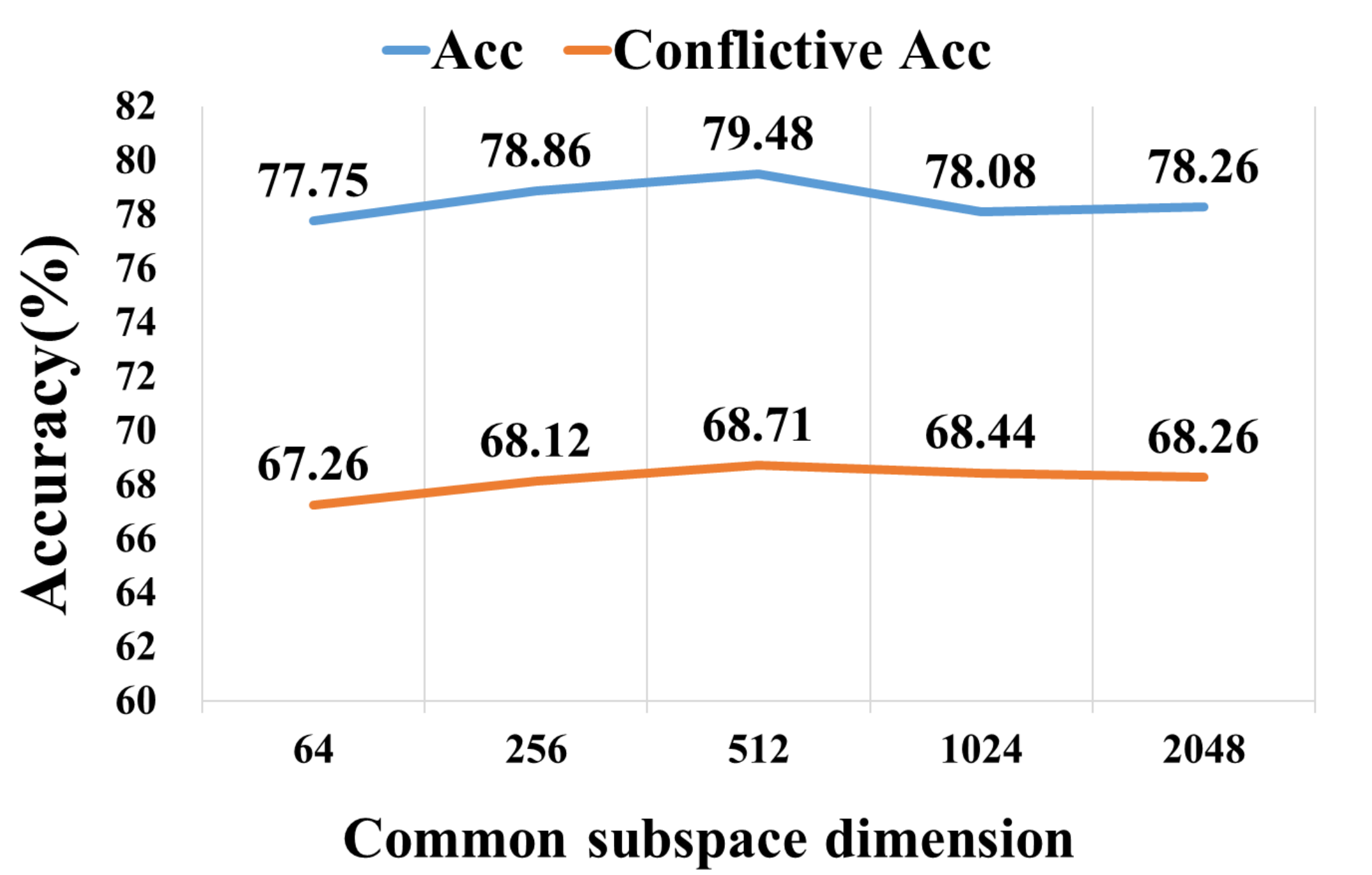} 
	\caption{Accuracy comparison with respect to $l$.}
	\label{dimension_study}
\end{figure}

\begin{table*}[h]
\caption{Accuracy (\%) on normal and conflictive test sets. The best results are highlighted in bold, while the second-best are underlined.}
\centering
\small
\begin{tabular}{cccccccc}
\hline
Datasets & Test sets & CPM-Nets & DUA-Nets & EDL & TMC & ECML & Ours \\ \hline
\multirow{2}{*}{HandWritten} & Normal & 94.60$\pm$0.57 & 97.30$\pm$0.77 & 97.05$\pm$0.50 & \underline{97.57$\pm$0.17} & 97.16$\pm$0.15 & \textbf{98.48 $\pm$ 0.12} \\
{} & Conflictive & 83.93$\pm$0.86 & 87.16$\pm$0.34 & 89.60$\pm$0.62 & 92.02$\pm$0.30 & \underline{92.69$\pm$0.20} & \textbf{94.65 $\pm$ 0.37} \\
\multirow{2}{*}{CUB} & Normal & 88.75$\pm$0.93 & 80.33$\pm$1.25 & 89.22$\pm$1.12 & 90.78$\pm$0.43 & \underline{92.00$\pm$0.37} & \textbf{92.48 $\pm$ 0.30} \\
{} & Conflictive & 68.55$\pm$1.10 & 54.92$\pm$4.12 & 70.79$\pm$0.91 & \underline{73.13$\pm$0.45} & \textbf{73.30$\pm$1.21} & 73.02 $\pm$ 0.96 \\
\multirow{2}{*}{Scene15} & Normal & 59.06$\pm$1.11 & 68.23$\pm$0.11 & 46.84$\pm$0.73 & 67.35$\pm$0.18 & \underline{70.80$\pm$0.16} & \textbf{77.75 $\pm$ 0.28} \\
{} & Conflictive & 28.25$\pm$1.48 & 27.45$\pm$1.90 & 36.68$\pm$0.72 & 42.92$\pm$0.35 & \underline{55.63$\pm$0.45} & \textbf{67.26 $\pm$ 0.43} \\
\multirow{2}{*}{PIE} & Normal & 88.68$\pm$0.82 & 82.66$\pm$2.02 & 82.64$\pm$0.94 & 91.91$\pm$0.43 & \textbf{94.11$\pm$0.68} & \underline{93.13 $\pm$ 0.50} \\
{} & Conflictive & 52.80$\pm$1.59 & 57.57$\pm$3.09 & 59.50$\pm$1.64 & 68.55$\pm$1.79 & \underline{81.94$\pm$0.66} & \textbf{83.13 $\pm$ 0.95} \\ 
\multirow{2}{*}{Caltech101} & Normal & 87.87$\pm$0.91 & 83.47$\pm$1.66 & 88.20$\pm$0.67 & 90.55$\pm$0.29 & \underline{91.13$\pm$1.61} & \textbf{93.61 $\pm$ 0.20} \\
{} & Conflictive & 64.56$\pm$0.51 & 78.47$\pm$1.92 & 84.14$\pm$0.34 & 87.25$\pm$0.34 & \underline{89.01$\pm$0.52} & \textbf{89.83 $\pm$ 0.31} \\
\multirow{2}{*}{NUS-WIDE} & Normal & 36.44$\pm$1.18 & 34.40$\pm$1.48 & 36.00$\pm$0.60 & 40.44$\pm$0.61 & \underline{42.97$\pm$0.54} & \textbf{47.72 $\pm$ 0.56} \\ 
{} & Conflictive & 24.98$\pm$1.07 & 31.81$\pm$2.41 & 31.79$\pm$0.58 & 33.25$\pm$0.57 & \underline{37.95$\pm$0.40} & \textbf{44.59 $\pm$ 0.37} \\
\multirow{2}{*}{NoisyMNIST} & Normal & 86.46$\pm$0.34 & 91.16$\pm$0.26 & 84.04$\pm$0.23 & \underline{90.93$\pm$0.05} & 90.88$\pm$0.08 & \textbf{96.17 $\pm$ 0.14} \\ 
 & Conflictive & 66.03$\pm$0.30 & 70.86$\pm$0.61 & 67.19$\pm$0.99 & \underline{71.42$\pm$0.04} & 71.39$\pm$0.04 & \textbf{72.33 $\pm$ 0.33} \\
\multirow{2}{*}{Event8} & Normal & 66.83$\pm$0.48 & 65.67$\pm$0.79 & 78.76$\pm$0.66 & 80.54$\pm$0.71 & \underline{82.51$\pm$0.27} & \textbf{84.14 $\pm$ 0.56} \\ 
 & Conflictive & 58.65$\pm$0.16 & 56.17$\pm$0.16 & 65.67$\pm$0.48 & 67.02$\pm$0.26 & \underline{68.22$\pm$0.26} & \textbf{70.60 $\pm$ 0.65} \\
\multirow{2}{*}{CCV} & Normal & 38.72$\pm$1.72 & 34.94$\pm$1.05 & 30.34$\pm$0.30 & 40.74$\pm$0.22 & \underline{41.88$\pm$0.06} & \textbf{49.69 $\pm$ 0.52} \\ 
 & Conflictive & 31.94$\pm$0.97 & 26.98$\pm$0.89 & 20.74$\pm$0.22 & 33.28$\pm$0.22 & \underline{36.06$\pm$0.29} & \textbf{43.98 $\pm$ 0.51} \\
\hline
\end{tabular}
\label{Table_Normal_Conflictive}
\end{table*}

To evaluate our model on extensive datasets, we consider both multi-view image datasets and feature-based datasets \cite{fang2023comprehensive}. Specifically, \textbf{HandWritten}\footnotemark \footnotetext{https://archive.ics.uci.edu/ml/datasets/Multiple+Features} comprises 2000 instances of handwritten numerals ranging form `0' to `9', with 200 patterns per class. It is represented using six feature sets. \textbf{CUB}\footnotemark \footnotetext{http://www.vision.caltech.edu/visipedia/CUB-200.html} consists of 11788 instances associated with text descriptions of 200 different categories of birds. In this paper, we focus on the first 10 categories and extract image features using GoogleNet and corresponding text features using doc2vec. \textbf{Scene15}\footnotemark \footnotetext{https://doi.org/10.6084/m9.figshare.7007177.v1} includes 4485 images from 15 indoor and outdoor scene categories. We extract three types of features GIST, PHOG, and LBP. \textbf{PIE}\footnotemark \footnotetext{http://www.cs.cmu.edu/afs/cs/project/PIE/MultiPie/MultiPie Home.html} contains 680 instances belonging to 68 classes. We extract intensity, LBP, and Gabor as 3 views. \textbf{Caltech101}\footnotemark \footnotetext{http://www.vision.caltech.edu/Image Datasets/Caltech101} comprises 8677 images from 101 classes.  We select 2386 samples from 20 classes, where each image has six features: Gabor, Wavelet Moments, CENTRIST, HOG, GIST, and LBP. \textbf{NUS-WIDE} \footnotemark \footnotetext{https://paperswithcode.com/dataset/nus-wide} consists of 269648 images with 81 concepts. For the top 12 classes, we select 200 images from each class, with a total of 6 different views. \textbf{NoisyMNIST}\footnotemark \footnotetext{https://www.kaggle.com/datasets/rahulnakka/noisymnist} contains 50000 training images. It is generated by introducing white Gaussian noise, motion blur and a combination of additive white Gaussian noise compared to the MNIST dataset. \textbf{Event8}\footnotemark \footnotetext{http://vision.stanford.edu/lijiali/event dataset} is a challenging dataset characterized by significant background variability. It comprises 8 sports event categories, 1579 samples, and three features. \textbf{CCV} \cite{jiang2011consumer} is a collection of YouTube videos encompassing 91 complex event categories, which provides extensive samples to facilitate training and benchmarking in complex video classification tasks. 
Table \ref{Dataset_summary} presents the details of datasets.  

\subsubsection{Compared Methods}

To convincingly demonstrate the advantages of our model, we compare it against state-of-the-art baselines in both feature fusion and decision fusion. Specifically, 1) feature fusion baselines include: \textbf{CPM-Nets} (Cross Partial Multi-view Networks) \cite{zhang2020deep} is a multi-view  feature fusion method designed to learn versatile representations that handle complex correlations across different views. \textbf{DUA-Nets} (Dynamic Uncertainty-Aware Networks) \cite{geng2021uncertainty} is an uncertainty-aware method,that uses reversal networks to integrate intrinsic information from different views into a unified representation. \textbf{EDL} (Evidential Deep Learning) \cite{sensoy2018evidential} designs the predictive distribution of the classification by placing a Dirichlet distribution on the class probabilities. 2) Decision fusion baselines include: \textbf{TMC} (Trusted Multi-view Classification) \cite{han2022trusted} addresses the uncertainty estimation problem and formulates a reliable classification framework. \textbf{ECML} (Evidential Conflictive Multi-view Learning) \cite{xu2024reliable} puts forward a conflictive opinion aggregation model.

\subsubsection{Implementation Details}
For all datasets, we randomly select 80\% of the samples for the training set and use the remaining 20\% for the test set. For our algorithm, the Adam optimizer is used to train the network, where $l_2$-norm regularization is set to $1e^{-5}$. We employ $5$-fold cross-validation to select the learning rate from $\{ 1e^{-4},3e^{-4},1e^{-3},3e^{-3} \}$. For selecting the dimension $l$ of the shared subspace of our model, we conduct accuracy comparison experiments with respect to dimension on the Scene15 dataset, while maintaining other parameters unchanged. The results on both normal and conflicting scenarios are shown in Fig. \ref{dimension_study}. It is seen that the model's accuracy initially increases with $l$ but eventually declines. While higher dimensions can enhance performance, they also significantly increase complexity. For example, the model's parameter file size grows from 158KB at $l=64$ to 52008KB at $l=2048$, reflecting a huge increase in complexity. Therefore, we choose $l=64$ for our model to make a trade-off between accuracy and computational cost. For other hyperparameters, both $\delta$ and $\gamma$ on all datasets are set to $1$. We fix $\delta$ to 1 based on the consideration that the common information shared across different views is typically dominant. Due to the diverse data features from different datasets, the parameter $\eta$ is selected from $\{0.1,0.01,0.001\}$. 

To generate a test set containing conflicting instances, we perform the following operations: (1) For noisy views, we add Gaussian noise with a fixed standard deviation $\sigma$ to $10\%$ of the test instances. (2) For unaligned views, we select $40\%$ of the test instances and modify the information in a random view, resulting in a misalignment between the label of that view and the true label of the instance. All methods are implemented by PyTorch on a NVIDIA GeForce RTX 4080 with GPU of 16GB memory. Experimental results are reported by averaging 10 independent trials.

%


\subsection{Experimental Results}

\subsubsection{Accuracy Comparison}

\begin{figure*}[h]
	\centering
	\includegraphics[scale=0.24]{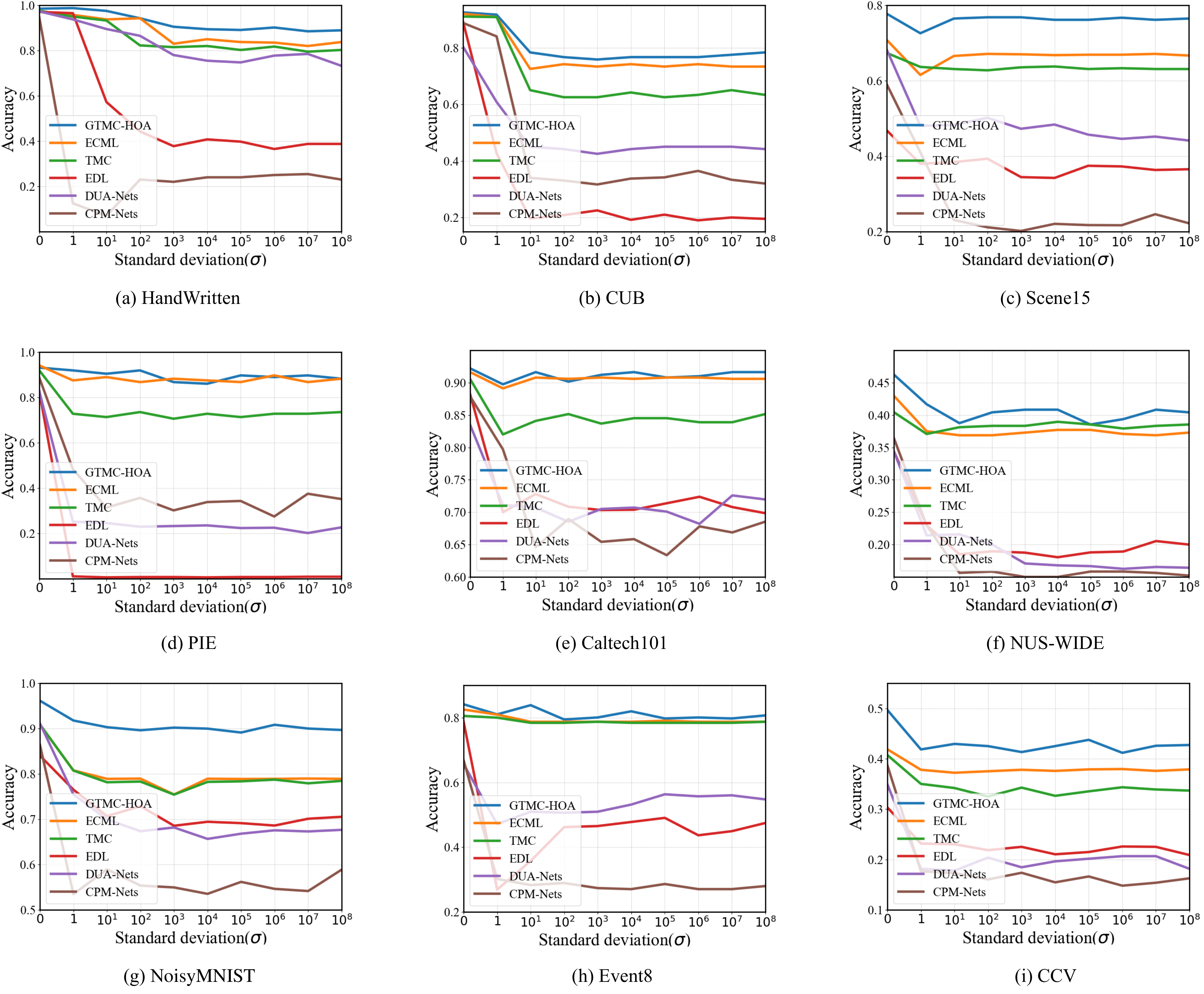} 
	\caption{Performance comparison on different datasets with varying noise levels.}
	\label{comparison_Varyingnoise}
\end{figure*}

Table \ref{Table_Normal_Conflictive} records the comparison results of various methods on normal and conflictive test sets. It is observed that CPM-Nets encounters a significant decline in performance on conflictive test sets compared to normal test sets, exemplified by a $35.88\%$ drop on the PIE dataset. This is because CPM-Nets cannot assess the quality of view data. In contrast, evidence theory-based methods demonstrate a certain ability to maintain performance on conflictive test sets, which is particularly evident for ECML and our GTMC-HOA. For instance, GTMC-HOA shows a decline of no more than $10\%$ on most datasets, except for the CUB dataset, where the decline reaches approximately $20\%$. This is attributed to CUB being a small-scale dataset with only two views, making it challenging for the model to conduct sufficient training to handle conflictive data. 

Importantly, we see that our proposed GTMC-HOA nearly outperforms all other baselines on both normal and conflictive test sets. For example, on normal test sets, GTMC-HOA outperforms other methods on the HandWritten, CUB, Scene15, Caltech101, and NUS-WIDE datasets. Notably, GTMC-HOA achieves a significant performance increase of $6.95\%$ on the Scene15 dataset compared to ECML, the second-best performing method. We argue that this is because the intra-view aggregation improves the view quality and reliability, which can be further supported in the uncertainty estimation study. In addition, we observe that even on large-scale datasets, our model consistently demonstrates significant advantages over existing baselines. For example, on the NoisyMNIST dataset, our GTMC-HOA approach achieves an accuracy of $96.17\%$, substantially outperforming TMC--the strongest competing baseline--which achieves only $90.93\%$. Moreover, on the CCV dataset, which is inherently derived from video data, GTMC-HOA continues to exhibit superior performance. Overall, extensive experiments across diverse and challenging datasets confirm the effectiveness and robustness of the proposed GTMC-HOA model.

\begin{figure}[htbp]
	\centering
	\includegraphics[scale=0.18]{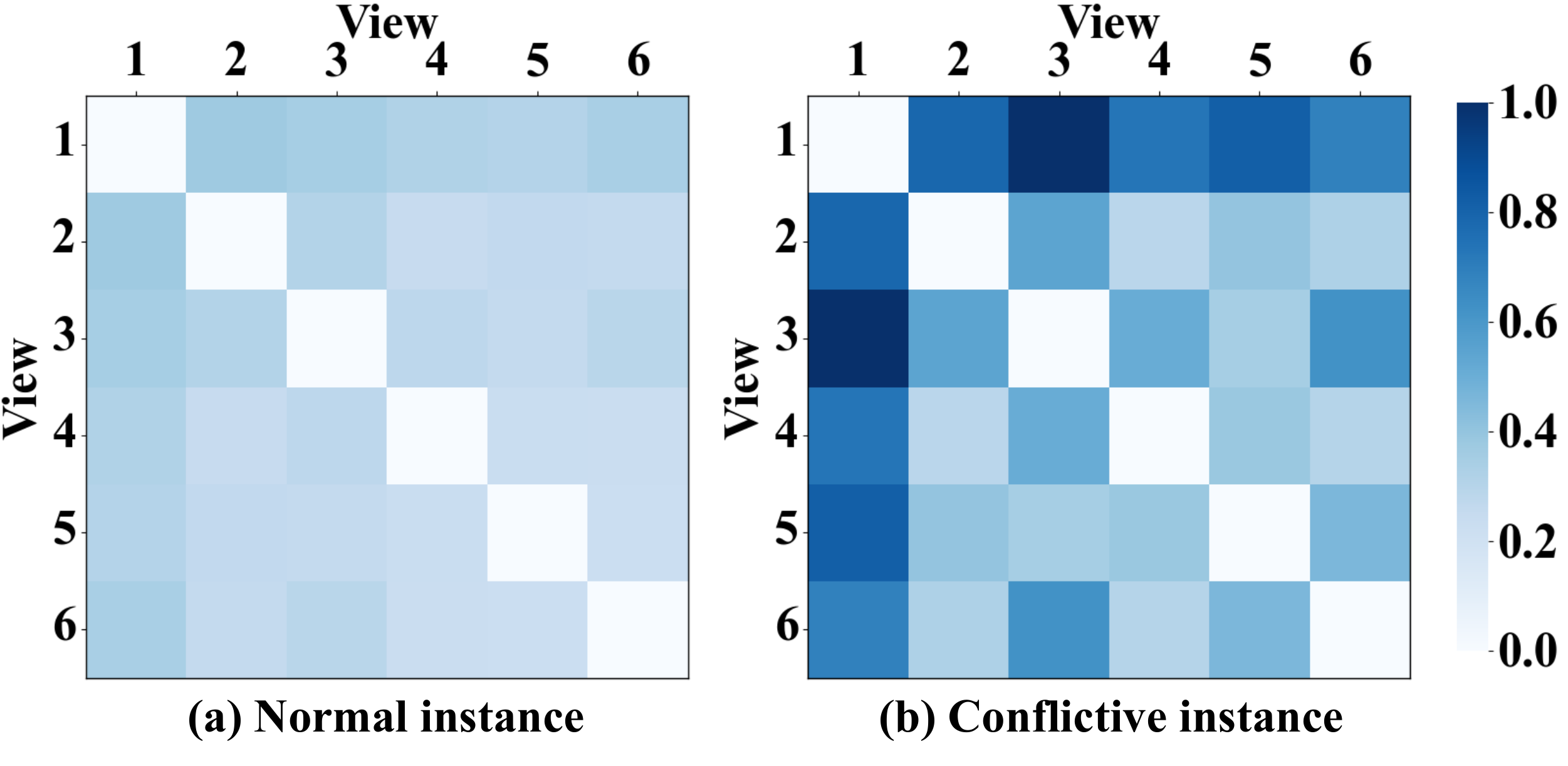} 
	\caption{Conflictive degree visualization. (a) conflictive degree of normal instances, (b) conflictive degree of conflictive instances.}
	\label{confVisualization}
\end{figure}

\begin{figure}[htbp]
	\centering
	\includegraphics[scale=0.15]{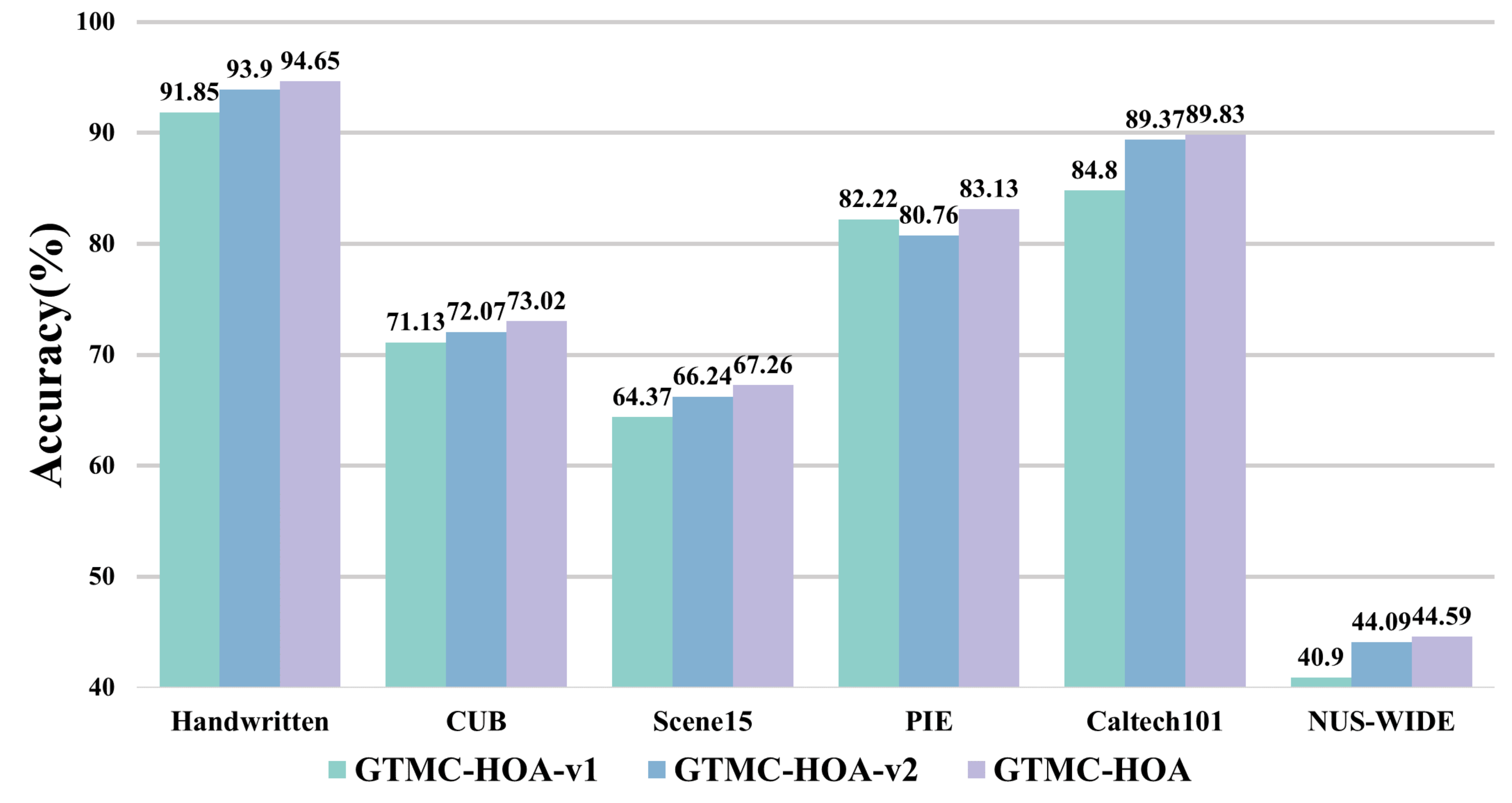} 
	\caption{Ablation study for evaluating the contributions of intra-view and inter-view aggregation.}
	\label{Ablation}
\end{figure} 

\begin{figure}[htbp]
	\centering
	\includegraphics[scale=0.15]{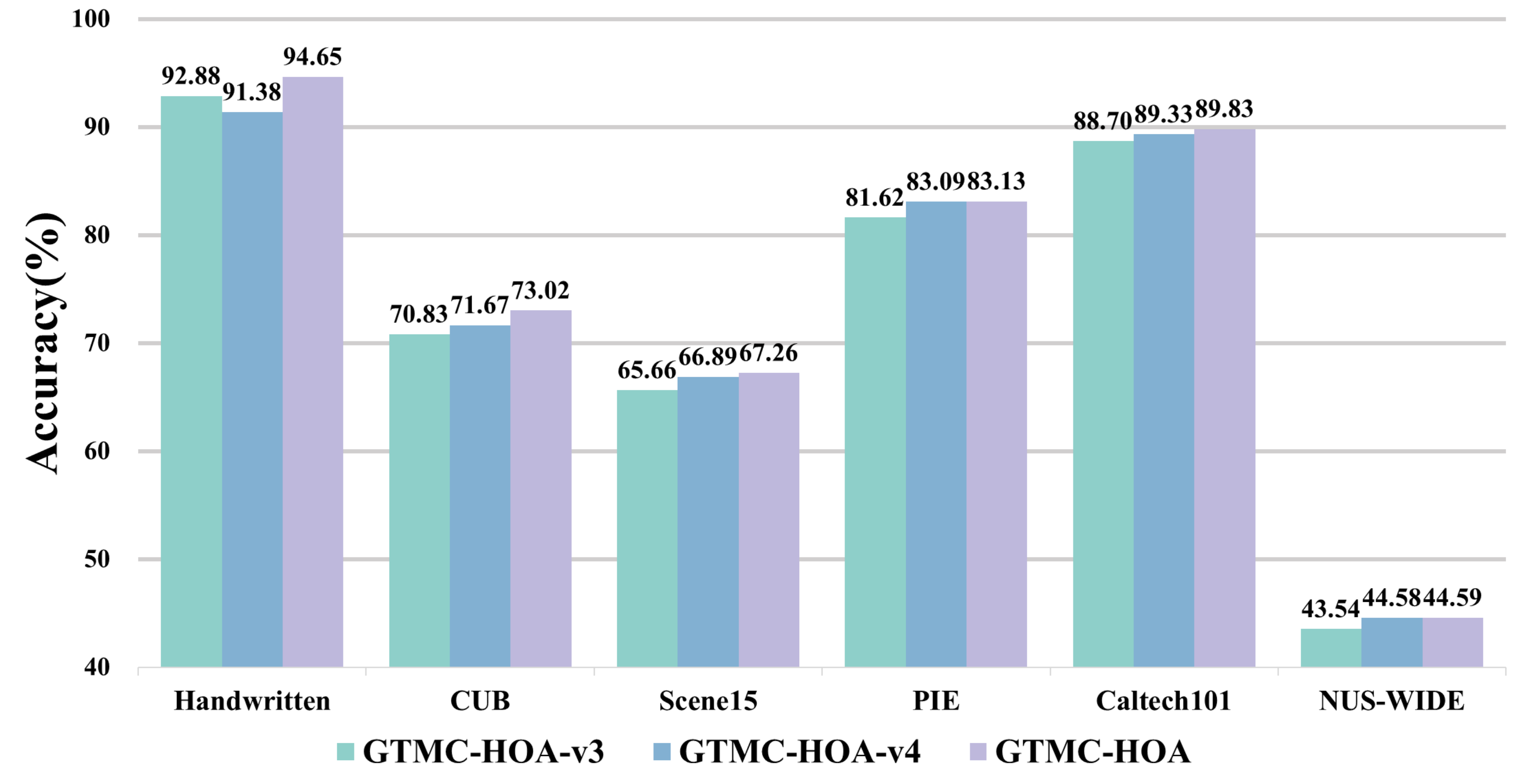} 
	\caption{Ablation study for evaluating the contributions of common and specific components.}
	\label{Ablation_exta}
\end{figure}  

\begin{figure}[htbp]
	\centering
	\includegraphics[scale=0.3]{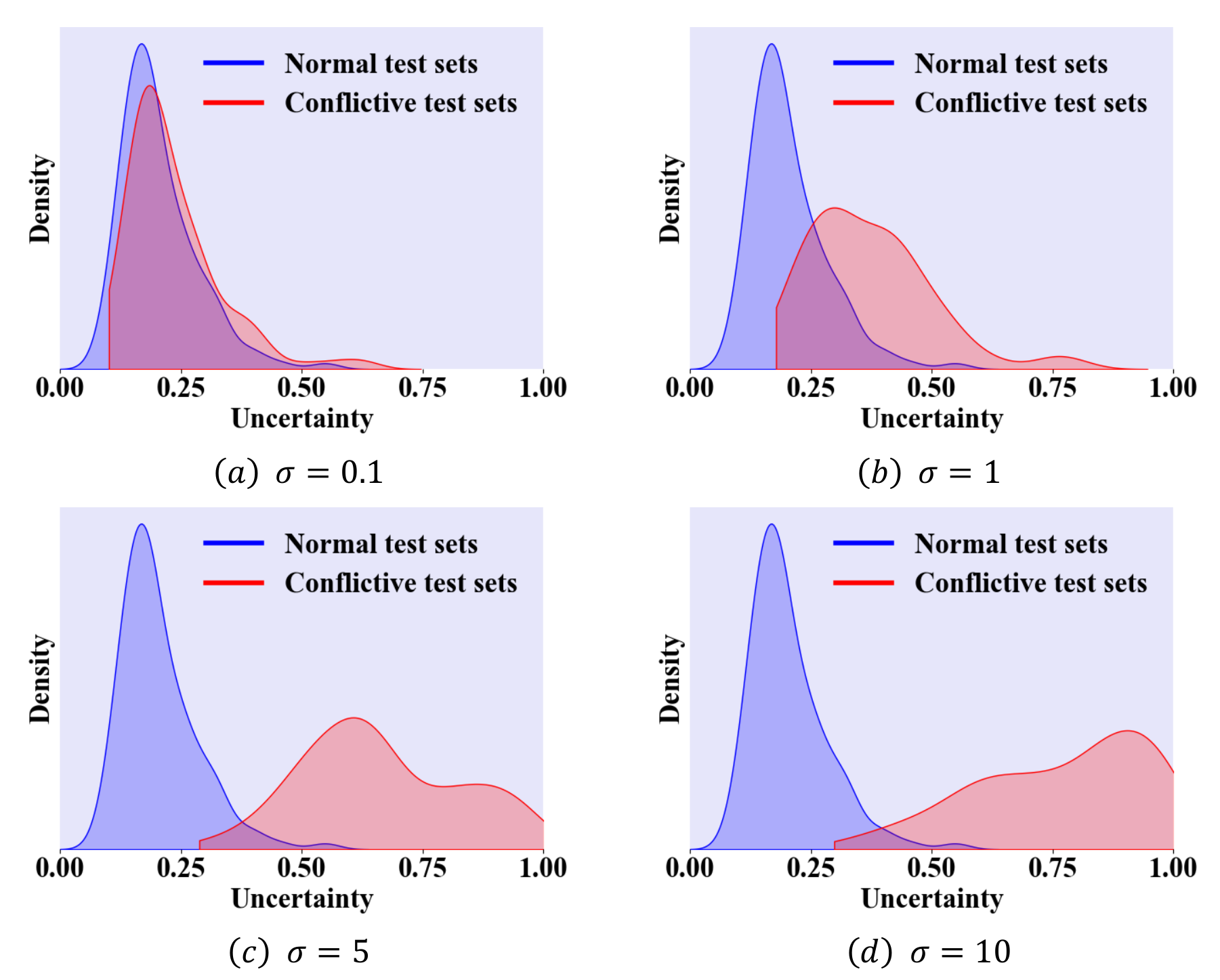} 
	\caption{Density of local uncertainty.}
	\label{denUncertainty_local}
\end{figure} 

\begin{figure}[htbp]
	\centering
	\includegraphics[scale=0.3]{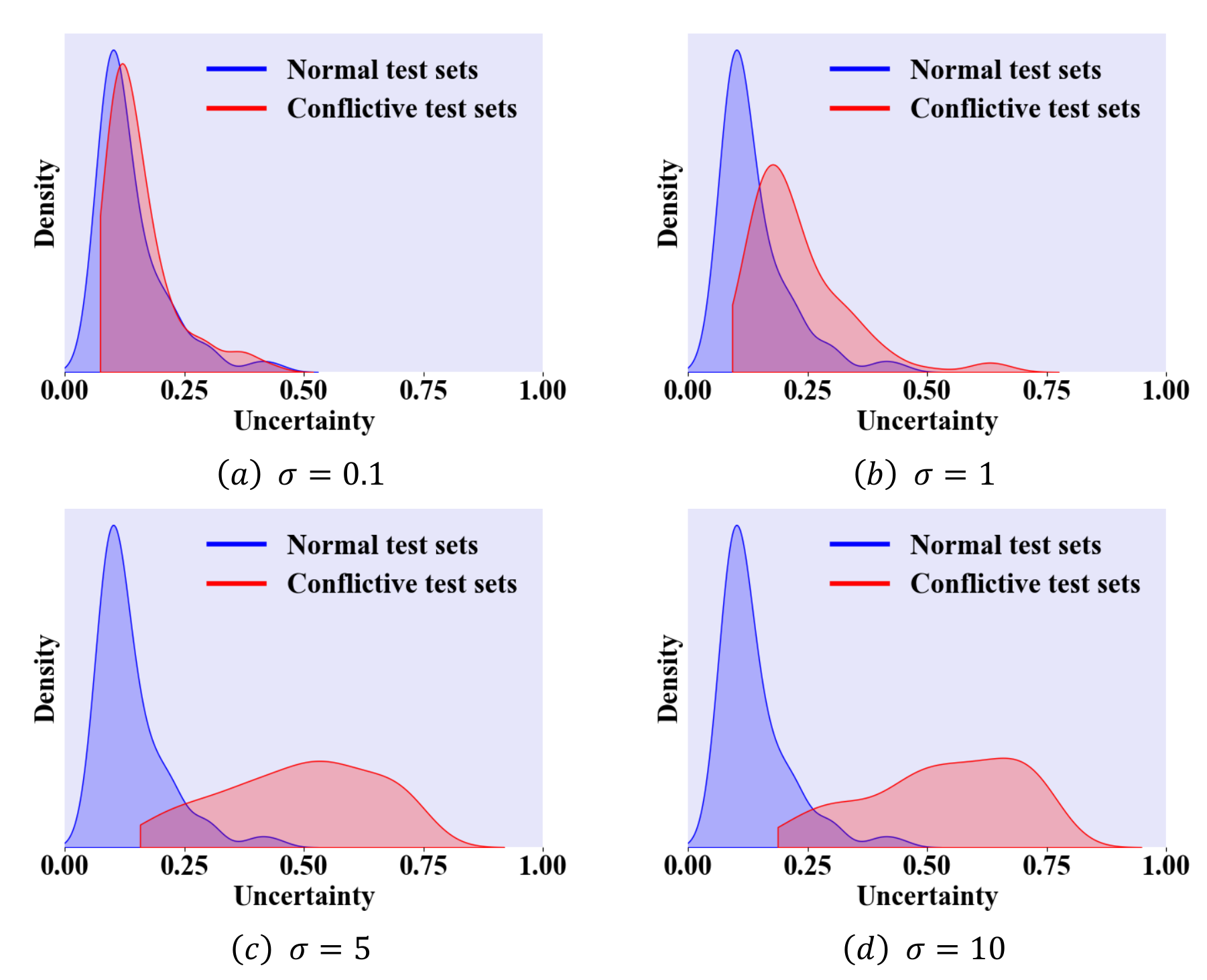} 
	\caption{Density of overall uncertainty.}
	\label{denUncertainty_whole}
\end{figure} 

\begin{figure*}[t]
   \centering
   \includegraphics[scale=0.3]{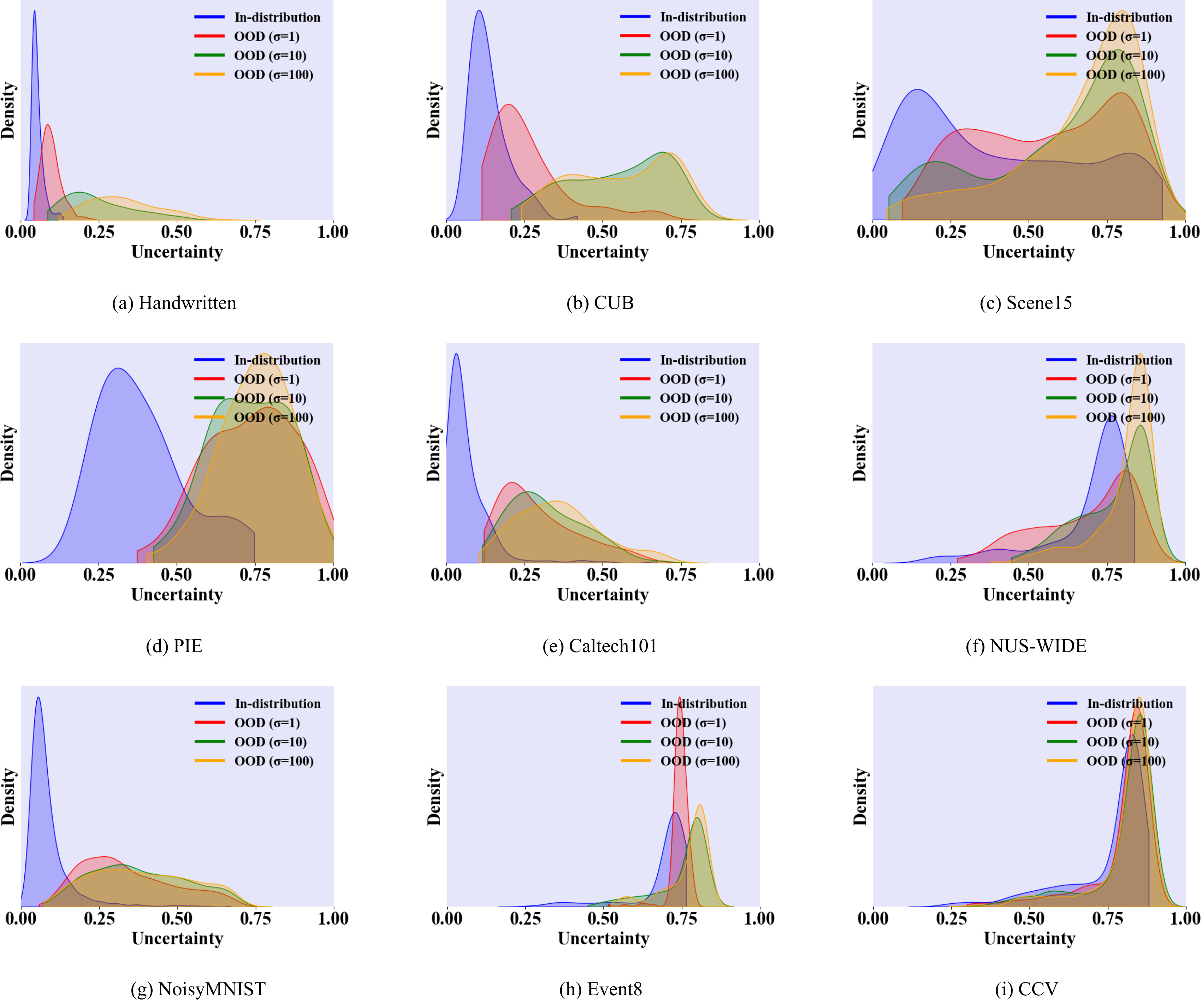} 
   \caption{Density of uncertainty obtained on different datasets.}
   \label{denUncertainty_AllDatasets}
\end{figure*}

\subsubsection{Performance Evaluation with Varying Noise Levels}

To further demonstrate the superiority of our GTMC-HOA model in the presence of noise, we conduct comparison experiments by adding Gaussian noise with different variances to the test data. Specifically, we randomly add Gaussian noise to approximately half of the views for all datasets, and the noise intensity, characterized by its standard deviation, ranges from $\{10^0, 10^1, \cdots, 10^8\}$. Fig. \ref{comparison_Varyingnoise} shows the comparison results on nine datasets with varying noise levels. It is observed that in the absence of Gaussian noise, CTMC-HOA outperforms other models. All tested models inevitably experience a decline after adding noise. It is worth noting that CMP-Nets and EDL exhibit a significant decline in performance in the presence of noise, especially when compared to TMC, ECML, and GTMC-HOA. This disparity arises because CMP-Nets is not based on evidence theory and thus cannot evaluate the uncertainty of noisy views, while EDL, being a single-view model, lacks the advantage of multi-view models that can mitigate noisy view limitations by leveraging information from other reliable views. Furthermore, we observe that model performance tends to stabilize at a certain accuracy level when noise intensity reaches a specific threshold. Importantly, compared to the latest trusted methods, namely TMC and ECML, our GTMC-HOA model demonstrates clear superiority across all tested datasets under the same noise level. This superiority is particularly pronounced on the Scene15, NoisyMNIST, and CCV datasets. The performance gains can be attributed to the proposed hierarchical aggregation framework, which effectively enhances uncertainty awareness during both intra- and inter-view opinion aggregation, thereby improving the reliability of decision-making.

\subsubsection{Conflictive Degree Visualization}
Fig. \ref{confVisualization} presents the conflictive degree result on the HandWritten dataset. To introduce conflicts, we modify the content in the first view, resulting in misalignment with the content from the other views. We observe and analyze the following: (1) The first view exhibits the highest level of conflict with other views, demonstrating that GTMC-HOA effectively captures conflict degrees among views. (2) The remaining view pairs, excluding those involving the first view, also show certain conflict degrees. This is because our model extracts common features across views, causing the conflictive information from one view to influence the others.


\subsubsection{Ablation Study}
We conduct ablation experiments to investigate the contributions of \emph{H1} and \emph{H2} hierarchy aggregations, as shown in Fig. \ref{Ablation}. In the first test, we remove the \emph{H1} hierarchy aggregation, and in the second, we exclude the attention mechanism of the \emph{H2} hierarchy aggregation. These two variants are denoted as GTMC-HOA-v1 and GTMC-HOA-v2, respectively. Our results indicate that both intra-view and inter-view aggregations contribute to overall performance. In particular, removing the first hierarchy of aggregation leads to a more significant performance decline, highlighting the importance of intra-view aggregation. 

In addition, we conduct more fine-grained ablation experiments to investigate the contributions of the common and specific components. Specifically, we remove the common information loss to obtain a variant denoted as GTMC-HOA-v3, and remove the view-specific information loss to obtain GTMC-HOA-v4. The corresponding results are shown in Fig. \ref{Ablation_exta}. It is seen that both variants GTMC-HOA-v3 and GTMC-HOA-v4 underperform compared to the original GTMC-HOA, indicating the importance of learning both common and specific information for optimal model performance. Additionally, except on the HandWritten dataset, GTMC-HOA-v3 shows a greater performance drop than GTMC-HOA-v4. This suggests that, compared to specific information, common information generally contributes more to predictive capability in decision-making, which aligns with conventional understanding.


\subsubsection{Uncertainty Estimation}

To evaluate the estimated uncertainty, we visualize the distribution of normal and conflictive test sets for both intra-view and inter-view aggregation. The distribution from intra-view aggregation represents local uncertainty, while the distribution from inter-view aggregation reflects overall uncertainty. Experiments are conducted on the CUB dataset, with conflictive test sets created by adding Gaussian noise with standard deviations of $\sigma=0.1,1,5,10$ to $50\%$ of the test instances, as shown in Figs. \ref{denUncertainty_local} and \ref{denUncertainty_whole}. It is observed that, in both cases, the distribution curves of conflictive instances deviate more from those of normal instances as noise intensity increases. Moreover, by comparing Fig. \ref{denUncertainty_local}(a) with \ref{denUncertainty_whole}(a), both under the same noise intensity of $\sigma$, we find that the distribution curve of overall uncertainty shifts to the left compared to that of local uncertainty. This shift implies a reduction in uncertainty after the second hierarchy of aggregation. 

In addition, we conduct more experiments regarding uncertainty estimation on all tested datasets, as shown in Fig. \ref{denUncertainty_AllDatasets}. We observe an evident shift between in-distribution and out-of-distribution performance on most datasets, such as Caltech101 and NoisyMNIST, consistent with the patterns described above. These observations indicate that both hierarchies of aggregation effectively estimate uncertainty, thereby improving view quality and overall reliability. 


\subsubsection{Practical Complexity}

To evaluate the practical computational and memory costs of our model, we calculate the number of learnable parameters, MACs, and FLOPs on different datasets, as shown in Table \ref{prac_metric}. The results reveal distinct patterns in model complexity influenced by the characteristics of each dataset. For instance, the model evaluated on the Handwritten dataset exhibits a compact architecture, with only 0.11M parameters and a per-sample computational cost of 0.017 GMACs, highlighting its efficiency for low-dimensional multi-view tasks. In contrast, the model tested on the Caltech101 dataset demonstrates the highest complexity, with 0.51M parameters and 0.116 GFLOPs. This increased complexity is attributed to the high dimensionality of the input features and the greater number of views. Notably, the Scene15 dataset—characterized by simpler view data—yields the most efficient configuration, requiring just 0.04M parameters and 0.012 GFLOPs. Overall, this analysis confirms that our model maintains good computational efficiency across diverse scenarios.

\begin{table}[htbp]
\centering
\caption{Model computational and parameter statistics across datasets}
\label{tab:model_stats}
\begin{tabular}{|c|c|c|c|}
\hline
Dataset & FLOPs (G) & MACs (G) & Params (M) \\
\hline
Handwritten & 0.034 & 0.017 & 0.11 \\
CUB         & 0.039 & 0.019 & 0.19 \\
Scene15     & 0.012 & 0.006  & 0.04 \\
PIE         & 0.044 & 0.022 & 0.18 \\
Caltech101     & 0.116 & 0.058 & 0.51 \\
NUS-WIDE    & 0.047 & 0.023 & 0.17 \\
\hline
\end{tabular}
\label{prac_metric}
\end{table}

\section{Conclusions}

In this paper, we proposed a hierarchical opinion aggregation framework for trusted multi-view classification, namely GTMC-HOA. GTMC-HOA is composed of two hierarchies of aggregation: intra-view and inter-view aggregation. The intra-view aggregation is realized by first learning common and view-specific information, and then aggregate them together based on a trusted strategy. This hierarchy of aggregation helps to reduce feature noise and improve view quality. The inter-view aggregation incorporates an evidence-level attention mechanism to determine attention scores among views, which guides more accurate opinion aggregation across views. Experimental results on six datasets confirmed the superiority of GTMC-HOA.

\bibliography{IEEE}

\begin{thebibliography}{10}
\providecommand{\url}[1]{#1}
\csname url@samestyle\endcsname
\providecommand{\newblock}{\relax}
\providecommand{\bibinfo}[2]{#2}
\providecommand{\BIBentrySTDinterwordspacing}{\spaceskip=0pt\relax}
\providecommand{\BIBentryALTinterwordstretchfactor}{4}
\providecommand{\BIBentryALTinterwordspacing}{\spaceskip=\fontdimen2\font plus
\BIBentryALTinterwordstretchfactor\fontdimen3\font minus
  \fontdimen4\font\relax}
\providecommand{\BIBforeignlanguage}[2]{{%
\expandafter\ifx\csname l@#1\endcsname\relax
\typeout{** WARNING: IEEEtran.bst: No hyphenation pattern has been}%
\typeout{** loaded for the language `#1'. Using the pattern for}%
\typeout{** the default language instead.}%
\else
\language=\csname l@#1\endcsname
\fi
#2}}
\providecommand{\BIBdecl}{\relax}
\BIBdecl

\bibitem{gao2015multi}
H.~Gao, F.~Nie, X.~Li, and H.~Huang, ``Multi-view subspace clustering,'' in
  \emph{Proceedings of the IEEE International Conference on Computer Vision},
  2015, pp. 4238--4246.

\bibitem{cao2023robust}
L.~Cao, L.~Shi, J.~Wang, Z.~Yang, and B.~Chen, ``Robust subspace clustering by
  logarithmic hyperbolic cosine function,'' \emph{IEEE Signal Processing
  Letters}, vol.~30, pp. 508--512, 2023.

\bibitem{wang2019gmc}
H.~Wang, Y.~Yang, and B.~Liu, ``Gmc: Graph-based multi-view clustering,''
  \emph{IEEE Transactions on Knowledge and Data Engineering}, vol.~32, no.~6,
  pp. 1116--1129, 2019.

\bibitem{wen2020adaptive}
J.~Wen, K.~Yan, Z.~Zhang, Y.~Xu, J.~Wang, L.~Fei, and B.~Zhang, ``Adaptive
  graph completion based incomplete multi-view clustering,'' \emph{IEEE
  Transactions on Multimedia}, vol.~23, pp. 2493--2504, 2020.

\bibitem{wang2020deep}
Q.~Wang, J.~Cheng, Q.~Gao, G.~Zhao, and L.~Jiao, ``Deep multi-view subspace
  clustering with unified and discriminative learning,'' \emph{IEEE
  Transactions on Multimedia}, vol.~23, pp. 3483--3493, 2020.

\bibitem{yan2021deep}
X.~Yan, S.~Hu, Y.~Mao, Y.~Ye, and H.~Yu, ``Deep multi-view learning methods: A
  review,'' \emph{Neurocomputing}, vol. 448, pp. 106--129, 2021.

\bibitem{kan2016multi}
M.~Kan, S.~Shan, and X.~Chen, ``Multi-view deep network for cross-view
  classification,'' in \emph{Proceedings of the IEEE Conference on Computer
  Vision and Pattern Recognition}, 2016, pp. 4847--4855.

\bibitem{wang2021fast}
S.~Wang, X.~Liu, X.~Zhu, P.~Zhang, Y.~Zhang, F.~Gao, and E.~Zhu, ``Fast
  parameter-free multi-view subspace clustering with consensus anchor
  guidance,'' \emph{IEEE Transactions on Image Processing}, vol.~31, pp.
  556--568, 2021.

\bibitem{lee2023mvfs}
Y.~Lee, Y.~Jeong, K.~Park, and S.~Kang, ``Mvfs: Multi-view feature selection
  for recommender system,'' in \emph{Proceedings of the 32nd ACM International
  Conference on Information and Knowledge Management}, 2023, pp. 4048--4052.

\bibitem{wang2015robust}
Y.~Wang, X.~Lin, L.~Wu, W.~Zhang, Q.~Zhang, and X.~Huang, ``Robust subspace
  clustering for multi-view data by exploiting correlation consensus,''
  \emph{IEEE Transactions on Image Processing}, vol.~24, no.~11, pp.
  3939--3949, 2015.

\bibitem{luo2018consistent}
S.~Luo, C.~Zhang, W.~Zhang, and X.~Cao, ``Consistent and specific multi-view
  subspace clustering,'' in \emph{Proceedings of the AAAI Conference on
  Artificial Intelligence}, vol.~32, no.~1, 2018.

\bibitem{han2022trusted}
Z.~Han, C.~Zhang, H.~Fu, and J.~T. Zhou, ``Trusted multi-view classification
  with dynamic evidential fusion,'' \emph{IEEE Transactions on Pattern Analysis
  and Machine Intelligence}, vol.~45, no.~2, pp. 2551--2566, 2022.

\bibitem{shafer1976mathematical}
G.~Shafer, \emph{A mathematical theory of evidence}.\hskip 1em plus 0.5em minus
  0.4em\relax Princeton University Press, 1976, vol.~42.

\bibitem{xu2024reliable}
C.~Xu, J.~Si, Z.~Guan, W.~Zhao, Y.~Wu, and X.~Gao, ``Reliable conflictive
  multi-view learning,'' in \emph{Proceedings of the AAAI Conference on
  Artificial Intelligence}, vol.~38, no.~14, 2024, pp. 16\,129--16\,137.

\bibitem{liu2022trusted}
W.~Liu, X.~Yue, Y.~Chen, and T.~Denoeux, ``Trusted multi-view deep learning
  with opinion aggregation,'' in \emph{Proceedings of the AAAI Conference on
  Artificial Intelligence}, vol.~36, no.~7, 2022, pp. 7585--7593.

\bibitem{xu2024trusted}
C.~Xu, Y.~Zhang, Z.~Guan, and W.~Zhao, ``Trusted multi-view learning with label
  noise,'' \emph{arXiv preprint arXiv:2404.11944}, 2024.

\bibitem{shi2024nonlinear}
L.~Shi, L.~Cao, Z.~Chen, Y.~Zhao, and B.~Chen, ``Nonlinear subspace clustering
  by functional link neural networks,'' \emph{Applied Soft Computing}, p.
  112303, 2024.

\bibitem{brbic2018multi}
M.~Brbi{\'c} and I.~Kopriva, ``Multi-view low-rank sparse subspace
  clustering,'' \emph{Pattern Recognition}, vol.~73, pp. 247--258, 2018.

\bibitem{zhang2018generalized}
C.~Zhang, H.~Fu, Q.~Hu, X.~Cao, Y.~Xie, D.~Tao, and D.~Xu, ``Generalized latent
  multi-view subspace clustering,'' \emph{IEEE Transactions on Pattern Analysis
  and Machine Intelligence}, vol.~42, no.~1, pp. 86--99, 2018.

\bibitem{chen2020multi}
M.-S. Chen, L.~Huang, C.-D. Wang, and D.~Huang, ``Multi-view clustering in
  latent embedding space,'' in \emph{Proceedings of the AAAI Conference on
  Artificial Intelligence}, vol.~34, no.~04, 2020, pp. 3513--3520.

\bibitem{shi2024enhanced}
L.~Shi, L.~Cao, J.~Wang, and B.~Chen, ``Enhanced latent multi-view subspace
  clustering,'' \emph{IEEE Transactions on Circuits and Systems for Video
  Technology}, 2024.

\bibitem{wu2024low}
T.~Wu, S.~Feng, and J.~Yuan, ``Low-rank kernel tensor learning for incomplete
  multi-view clustering,'' in \emph{Proceedings of the AAAI Conference on
  Artificial Intelligence}, vol.~38, no.~14, 2024, pp. 15\,952--15\,960.

\bibitem{liu2019multiple}
X.~Liu, X.~Zhu, M.~Li, L.~Wang, E.~Zhu, T.~Liu, M.~Kloft, D.~Shen, J.~Yin, and
  W.~Gao, ``Multiple kernel $ k $ k-means with incomplete kernels,'' \emph{IEEE
  Transactions on Pattern Analysis and Machine Intelligence}, vol.~42, no.~5,
  pp. 1191--1204, 2019.

\bibitem{liu2020optimal}
J.~Liu, X.~Liu, J.~Xiong, Q.~Liao, S.~Zhou, S.~Wang, and Y.~Yang, ``Optimal
  neighborhood multiple kernel clustering with adaptive local kernels,''
  \emph{IEEE Transactions on Knowledge and Data Engineering}, vol.~34, no.~6,
  pp. 2872--2885, 2020.

\bibitem{xie2018unifying}
Y.~Xie, D.~Tao, W.~Zhang, Y.~Liu, L.~Zhang, and Y.~Qu, ``On unifying multi-view
  self-representations for clustering by tensor multi-rank minimization,''
  \emph{International Journal of Computer Vision}, vol. 126, pp. 1157--1179,
  2018.

\bibitem{wu2020unified}
J.~Wu, X.~Xie, L.~Nie, Z.~Lin, and H.~Zha, ``Unified graph and low-rank tensor
  learning for multi-view clustering,'' in \emph{Proceedings of the AAAI
  Conference on Artificial intelligence}, vol.~34, no.~04, 2020, pp.
  6388--6395.

\bibitem{qin2023flexible}
Y.~Qin, Z.~Tang, H.~Wu, and G.~Feng, ``Flexible tensor learning for multi-view
  clustering with markov chain,'' \emph{IEEE Transactions on Knowledge and Data
  Engineering}, 2023.

\bibitem{zhan2018multiview}
K.~Zhan, F.~Nie, J.~Wang, and Y.~Yang, ``Multiview consensus graph
  clustering,'' \emph{IEEE Transactions on Image Processing}, vol.~28, no.~3,
  pp. 1261--1270, 2018.

\bibitem{li2021consensus}
Z.~Li, C.~Tang, X.~Liu, X.~Zheng, W.~Zhang, and E.~Zhu, ``Consensus graph
  learning for multi-view clustering,'' \emph{IEEE Transactions on Multimedia},
  vol.~24, pp. 2461--2472, 2021.

\bibitem{wang2022towards}
H.~Wang, G.~Jiang, J.~Peng, R.~Deng, and X.~Fu, ``Towards adaptive consensus
  graph: multi-view clustering via graph collaboration,'' \emph{IEEE
  Transactions on Multimedia}, vol.~25, pp. 6629--6641, 2022.

\bibitem{wong2019clustering}
W.~K. Wong, N.~Han, X.~Fang, S.~Zhan, and J.~Wen, ``Clustering
  structure-induced robust multi-view graph recovery,'' \emph{IEEE Transactions
  on Circuits and Systems for Video Technology}, vol.~30, no.~10, pp.
  3584--3597, 2019.

\bibitem{lin2021multi}
Z.~Lin, Z.~Kang, L.~Zhang, and L.~Tian, ``Multi-view attributed graph
  clustering,'' \emph{IEEE Transactions on Knowledge and Data Engineering},
  vol.~35, no.~2, pp. 1872--1880, 2021.

\bibitem{lin2021graph}
Z.~Lin and Z.~Kang, ``Graph filter-based multi-view attributed graph
  clustering.'' in \emph{IJCAI}, 2021, pp. 2723--2729.

\bibitem{chen2023deep}
J.~Chen, H.~Mao, W.~L. Woo, and X.~Peng, ``Deep multiview clustering by
  contrasting cluster assignments,'' in \emph{Proceedings of the IEEE/CVF
  International Conference on Computer Vision}, 2023, pp. 16\,752--16\,761.

\bibitem{lu2024decoupled}
Y.~Lu, Y.~Lin, M.~Yang, D.~Peng, P.~Hu, and X.~Peng, ``Decoupled contrastive
  multi-view clustering with high-order random walks,'' in \emph{Proceedings of
  the AAAI Conference on Artificial Intelligence}, vol.~38, no.~13, 2024, pp.
  14\,193--14\,201.

\bibitem{zhu2024multiview}
P.~Zhu, X.~Yao, Y.~Wang, B.~Hui, D.~Du, and Q.~Hu, ``Multiview deep subspace
  clustering networks,'' \emph{IEEE Transactions on Cybernetics}, 2024.

\bibitem{liu2023incomplete}
C.~Liu, J.~Wen, X.~Luo, and Y.~Xu, ``Incomplete multi-view multi-label learning
  via label-guided masked view-and category-aware transformers,'' in
  \emph{Proceedings of the AAAI Conference on Artificial Intelligence},
  vol.~37, no.~7, 2023, pp. 8816--8824.

\bibitem{zhu2022latent}
P.~Zhu, X.~Yao, Y.~Wang, M.~Cao, B.~Hui, S.~Zhao, and Q.~Hu, ``Latent
  heterogeneous graph network for incomplete multi-view learning,'' \emph{IEEE
  Transactions on Multimedia}, vol.~25, pp. 3033--3045, 2022.

\bibitem{wang2024integrated}
Y.~Wang, X.~Yao, P.~Zhu, W.~Li, M.~Cao, and Q.~Hu, ``Integrated heterogeneous
  graph attention network for incomplete multi-modal clustering,''
  \emph{International Journal of Computer Vision}, vol. 132, no.~9, pp.
  3847--3866, 2024.

\bibitem{yang2024fast}
B.~Yang, X.~Zhang, J.~Wu, F.~Nie, Z.~Lin, F.~Wang, and B.~Chen, ``Fast
  multiview anchor-graph clustering,'' \emph{IEEE Transactions on Neural
  Networks and Learning Systems}, 2024.

\bibitem{sensoy2018evidential}
M.~Sensoy, L.~Kaplan, and M.~Kandemir, ``Evidential deep learning to quantify
  classification uncertainty,'' \emph{Advances in Neural Information Processing
  Systems}, vol.~31, 2018.

\bibitem{han2020trusted}
Z.~Han, C.~Zhang, H.~Fu, and J.~T. Zhou, ``Trusted multi-view classification,''
  in \emph{International Conference on Learning Representations}, 2020.

\bibitem{liu2024dynamic}
Y.~Liu, L.~Liu, C.~Xu, X.~Song, Z.~Guan, and W.~Zhao, ``Dynamic evidence
  decoupling for trusted multi-view learning,'' in \emph{ACM Multimedia 2024}.

\bibitem{zhou2023calm}
H.~Zhou, Z.~Xue, Y.~Liu, B.~Li, J.~Du, M.~Liang, and Y.~Qi, ``Calm: An enhanced
  encoding and confidence evaluating framework for trustworthy multi-view
  learning,'' in \emph{Proceedings of the 31st ACM International Conference on
  Multimedia}, 2023, pp. 3108--3116.

\bibitem{zhou2023rtmc}
H.~Zhou, Z.~Xue, Y.~Liu, B.~Li, J.~Du, and M.~Liang, ``Rtmc: A rubost trusted
  multi-view classification framework,'' in \emph{2023 IEEE International
  Conference on Multimedia and Expo (ICME)}.\hskip 1em plus 0.5em minus
  0.4em\relax IEEE, 2023, pp. 576--581.

\bibitem{wang2024trusted}
X.~Wang, Y.~Wang, Y.~Wang, A.~Huang, and J.~Liu, ``Trusted semi-supervised
  multi-view classification with contrastive learning,'' \emph{IEEE
  Transactions on Multimedia}, 2024.

\bibitem{du2023bridging}
S.~Du, Z.~Fang, S.~Lan, Y.~Tan, M.~G{\"u}nther, S.~Wang, and W.~Guo, ``Bridging
  trustworthiness and open-world learning: An exploratory neural approach for
  enhancing interpretability, generalization, and robustness,'' in
  \emph{Proceedings of the 31st ACM International Conference on Multimedia},
  2023, pp. 8719--8729.

\bibitem{zou2023dpnet}
X.~Zou, C.~Tang, X.~Zheng, Z.~Li, X.~He, S.~An, and X.~Liu, ``Dpnet: Dynamic
  poly-attention network for trustworthy multi-modal classification,'' in
  \emph{Proceedings of the 31st ACM International Conference on Multimedia},
  2023, pp. 3550--3559.

\bibitem{liu2017adversarial}
P.~Liu, X.~Qiu, and X.-J. Huang, ``Adversarial multi-task learning for text
  classification,'' in \emph{Proceedings of the 55th Annual Meeting of the
  Association for Computational Linguistics (Volume 1: Long Papers)}, 2017, pp.
  1--10.

\bibitem{wu2019multi}
X.~Wu, Q.-G. Chen, Y.~Hu, D.~Wang, X.~Chang, X.~Wang, and M.-L. Zhang,
  ``Multi-view multi-label learning with view-specific information
  extraction.'' in \emph{IJCAI}, 2019, pp. 3884--3890.

\bibitem{jsang2018subjective}
A.~Jsang, \emph{Subjective Logic: A formalism for reasoning under
  uncertainty}.\hskip 1em plus 0.5em minus 0.4em\relax Springer Publishing
  Company, Incorporated, 2018.

\bibitem{fang2023comprehensive}
U.~Fang, M.~Li, J.~Li, L.~Gao, T.~Jia, and Y.~Zhang, ``A comprehensive survey
  on multi-view clustering,'' \emph{IEEE Transactions on Knowledge and Data
  Engineering}, vol.~35, no.~12, pp. 12\,350--12\,368, 2023.

\bibitem{jiang2011consumer}
Y.-G. Jiang, G.~Ye, S.-F. Chang, D.~Ellis, and A.~C. Loui, ``Consumer video
  understanding: A benchmark database and an evaluation of human and machine
  performance,'' in \emph{Proceedings of the 1st ACM international conference
  on multimedia retrieval}, 2011, pp. 1--8.

\bibitem{zhang2020deep}
C.~Zhang, Y.~Cui, Z.~Han, J.~T. Zhou, H.~Fu, and Q.~Hu, ``Deep partial
  multi-view learning,'' \emph{IEEE Transactions on Pattern Analysis and
  Machine Intelligence}, vol.~44, no.~5, pp. 2402--2415, 2020.

\bibitem{geng2021uncertainty}
Y.~Geng, Z.~Han, C.~Zhang, and Q.~Hu, ``Uncertainty-aware multi-view
  representation learning,'' in \emph{Proceedings of the AAAI Conference on
  Artificial Intelligence}, vol.~35, no.~9, 2021, pp. 7545--7553.

\end{thebibliography}

\bibliographystyle{IEEEtran}

\begin{IEEEbiography}[{\includegraphics[width=1in,height=1.25in,clip,keepaspectratio]{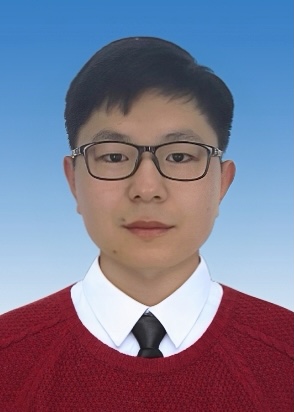}}]{Long Shi} (Member, IEEE) was born in Jiangsu Province, China. He received the Ph.D degree in electrical engineering from Southwest Jiaotong University, Chengdu, China, 2020. From 2018 to 2019, he was a visiting student with the Department of Electronic Engineering, University of York, U.K. Since 2021, he has been with the School of Computing and Artificial Intelligence, Southwestern University of Finance and Economics, where he is currently an associate professor. He has published several high-quality papers in IEEE TSP, IEEE TMM, IEEE TCSVT, IEEE SPL, IEEE TCSII, etc. His research interest lies at the intersection of signal processing and machine learning, including adaptive signal processing, multi-view (multimodal) learning, large language model.\end{IEEEbiography}

\begin{IEEEbiography}[{\includegraphics[width=1in,height=1.25in,clip,keepaspectratio]{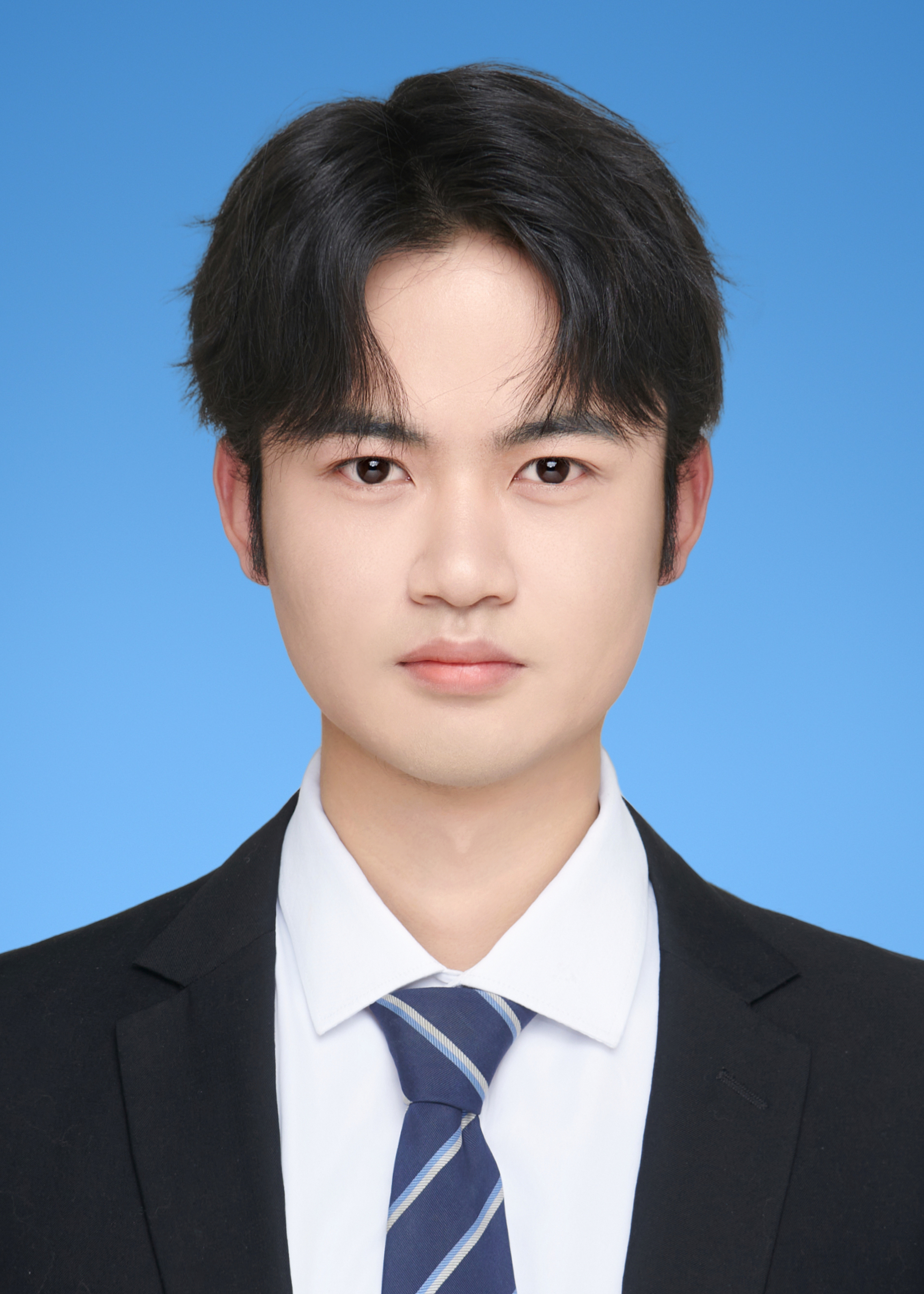}}]{Chuanqing Tang} is currently working toward a Master's degree at the School of Computing and Artificial Intelligence, Southwestern University of Finance and Economics (SWUFE), and is affiliated with the Financial Intelligence and Financial Engineering Key Laboratory of Sichuan Province (FlFE). Supervised by Dr. Long Shi, his research focuses on multi-view learning, multimodal machine learning, and trustworthy artificial intelligence..\end{IEEEbiography}

\begin{IEEEbiography}[{\includegraphics[width=1in,height=1.25in,clip,keepaspectratio]{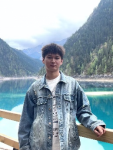}}]{Huangyi Deng} is currently pursuing a Master's degree at School of Computing and Artificial Intelligence, Financial Intelligence and Financial Engineering Key Laboratory of Sichuan Province (FlFE), Southwestern University of Finance and Economics(SWUEE), supervised by Dr. Long Shi. His research interestsare mainly in incomplete multi-view classification, trustworthy multi-view learning.\end{IEEEbiography}

\begin{IEEEbiography}[{\includegraphics[width=1in,height=1.25in,clip,keepaspectratio]{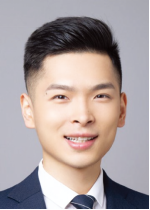}}]{Cai Xu} is an associate professor in the School of Computer Science and Technology at Xidian University. He has authored 34 papers published in renowned IEEE journals and esteemed academic conferences. His research interests include trustworthy machine learning, multi-view learning, multimedia understanding and recommender systems.\end{IEEEbiography}

\begin{IEEEbiography}[{\includegraphics[width=1in,height=1.25in,clip,keepaspectratio]{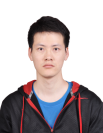}}]{Lei Xing} received the B.S. and Ph.D. degrees in control theory and engineering from Xi’an Jiaotong University, in 2014 and 2021, respectively. He was a Postdoctoral Researcher with the Hong Kong Polytechnic University from 2022 to 2023. Currently, he serves as an assistant professor at the Institute of Artificial Intelligence and Robotics (IAIR), Xi’an Jiaotong University. His research interests are in robust machine learning, multi-view learning and artificial intelligence. Dr. Xing has served as a reviewer for various journals or conferences.\end{IEEEbiography}

\begin{IEEEbiography}[{\includegraphics[width=1in,height=1.25in,clip,keepaspectratio]{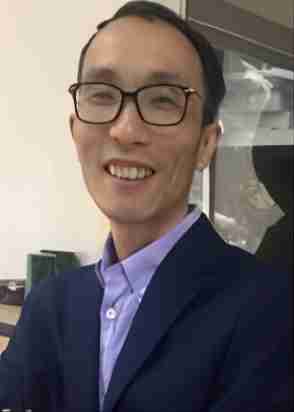}}]{Badong Chen} received the Ph.D. degree in Computer Science and Technology from Tsinghua University, Beijing, China, in 2008. He is currently a professor with the Institute of Artificial Intelligence and Robotics, Xi'an Jiaotong University, Xi'an, China. His research interests are in signal processing, machine learning, artificial intelligence and robotics. He has authored or coauthored over 300 articles in various journals and conference proceedings (with 13000+ citations in Google Scholar), and has won the 2022 Outstanding Paper Award of IEEE Transactions on Cognitive and Developmental Systems. Dr. Chen serves as a Member of the Machine Learning for Signal Processing Technical Committee of the IEEE Signal Processing Society, and serves (or has served) as an Associate Editor for several journals including IEEE Transactions on Neural Networks and Learning Systems, IEEE Transactions on Cognitive and Developmental Systems, IEEE Transactions on Circuits and Systems for Video Technology, Neural Networks and Journal of The Franklin Institute. He has served as a PC or SPC Member for prestigious conferences including UAI, IJCAI and AAAI, and served as a General Co-Chair of 2022 IEEE International Workshop on Machine Learning for Signal Processing.\end{IEEEbiography}










\vspace{11pt}

\vfill

\end{document}